\documentclass{article}

\usepackage[utf8]{inputenc} 
\usepackage[T1]{fontenc}    
\usepackage{hyperref}       
\usepackage{url}            
\usepackage{booktabs}       
\usepackage{amsfonts}       
\usepackage{nicefrac}       
\usepackage{microtype}      
\usepackage{xcolor}         
\usepackage{graphicx}
\usepackage{amsmath}
\usepackage{threeparttable}
\usepackage{multirow}
\usepackage{fontawesome5}
\usepackage{caption}
\usepackage[table]{xcolor}
\usepackage{wrapfig}
\usepackage{subcaption}
\usepackage{makecell}
\usepackage{algorithm}
\usepackage{algpseudocode}
\usepackage{amssymb}
\usepackage[main, final]{neurips_2026}
\definecolor{mycol1}{RGB}{200, 235, 210}
\definecolor{mycol2}{RGB}{211, 245, 225}
\definecolor{mycol3}{RGB}{211, 211, 211}
\makeatletter
\newcommand\latefnote[2]{%
  \protected@xdef\@thanks{\@thanks\protect\footnotetext[#1]{#2}}}
\makeatother

\title{
OmniSelect: Dynamic Modality-Aware Token Compression for Efficient Omni-modal Large Language Models
}

\author{%
\NoHyper
\begin{tabular}{c}
Morunliu Yang$^{1,3}$\thanks{Equal Contribution},\quad
Ruotao Xu$^1$\footnotemark[1],\quad
Le Li$^1$,\quad Yue Wang$^1$, \quad Jianxin Zhang$^1$ ,\\
\quad Yihang Lou$^2$,\quad Siwei Feng$^{1}$,\quad
Peifeng Li$^{1}$ \quad Juntao Li$^{1,3}$
\thanks{Corresponding author}
\latefnote{3}{Part of this work was completed when Morunliu Yang interned at Meta-Stone (Pinghu) Co., Ltd.}
\end{tabular}
\and
\begin{tabular}{c}
$^{1}$Soochow University \quad $^{2}$Peking University
\quad $^{3}$Meta-Stone (Pinghu) Co., Ltd.
\end{tabular}
\and
\begin{tabular}{c}
\texttt{\{mrlyangnlp, 20255227018\}@stu.suda.edu.cn}
\quad
\texttt{\{ljt\}@suda.edu.cn}
\end{tabular}
\endNoHyper
}

\begin{document}

\maketitle

\vspace{-2em}
\begin{center}
    \textbf{\texttt{\faGithub~Code: \textcolor{violet}{ \url{https://github.com/Yangmrl-nlp/OmniSelect}}}}
\end{center}
\vspace{-0.5em}

\begin{figure*}[!hbtp]
    \centering
    \includegraphics[width=1.0\textwidth]{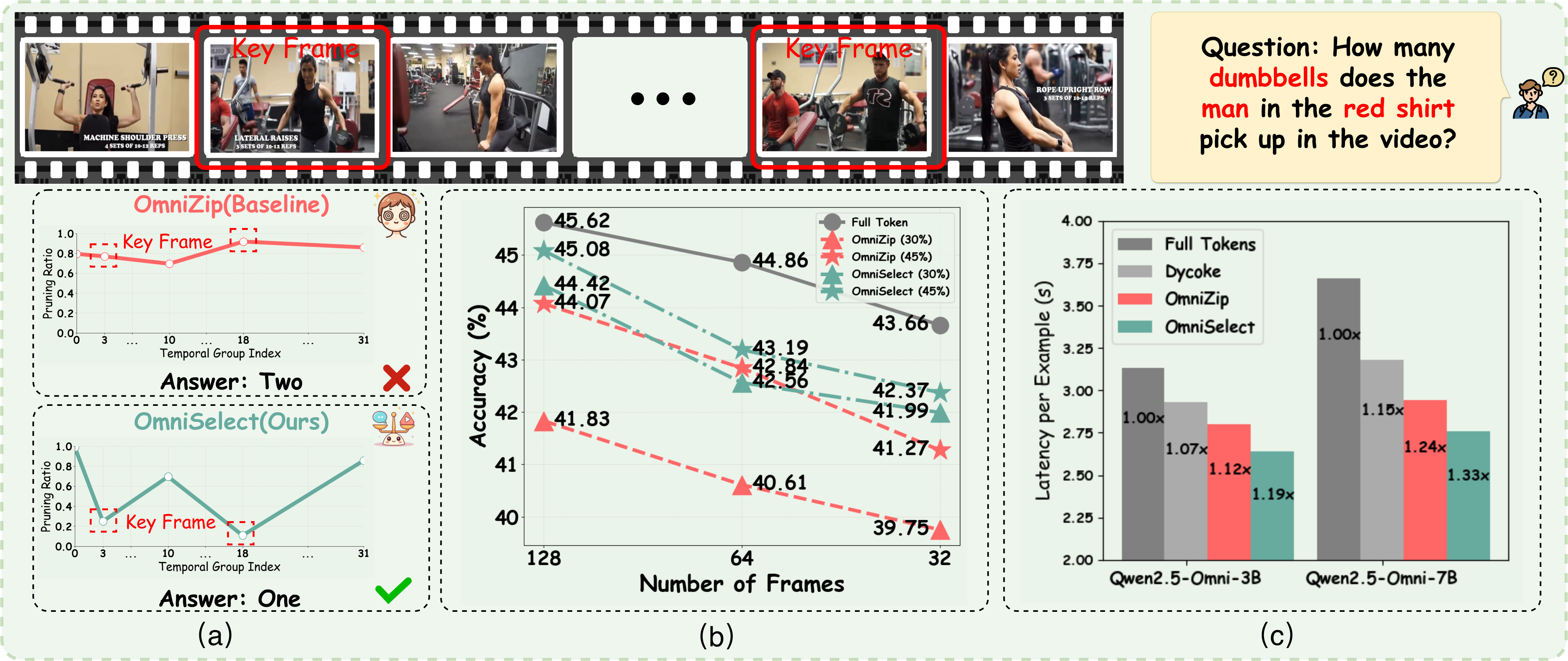}
    \caption{\textbf{(a)}: OmniSelect prunes fewer tokens in key frames and more in less important ones, while OmniZip prunes uniformly.
\textbf{(b)}: OmniSelect retains 94\% to 99\% of the original full-token accuracy (Qwen2.5-Omni-3B, Worldsense, 128 Frames) and achieves competitive performance among existing training-free approaches.\textbf{(c)}: OmniSelect achieves an inference speedup of 1.19$\times$ to 1.33$\times$.}
    \label{fig:overview}
\end{figure*}

\begin{abstract}

Omnimodal large language models (OmniLLMs) have recently gained increasing attention for unified audio-video understanding. However, processing long multimodal token sequences introduces substantial computational overhead, making efficient token compression crucial. Existing methods typically rely on fixed, modality-specific guidance, which fails to account for the varying importance of modalities across different queries.
To address this limitation, we propose \textbf{OmniSelect}, a training-free, modality-adaptive token pruning framework that dynamically selects appropriate compression strategies for multimodal inputs. Specifically, we leverage a lightweight AudioCLIP model to estimate cross-modal relevance and categorize each input into three pruning regimes: Audio-Centric, Video-Centric, and Uniform pruning. Based on these relevance scores, OmniSelect further performs fine-grained token pruning within each temporal group, adaptively allocating pruning ratios to preserve informative tokens across modalities.
By explicitly modeling modality preference and enabling dynamic strategy selection, OmniSelect effectively avoids the pitfalls of one-size-fits-all compression. Extensive experiments demonstrate that our method achieves efficient multimodal token reduction while maintaining strong performance, without requiring any additional training.

\end{abstract}

\section{Introduction} 

Conventional Video Large Language Models (VLLMs) have achieved remarkable success in video question-answering and comprehension \citep{bai2025qwen25vltechnicalreport, chen2024internvlscalingvisionfoundation, liu2023visualinstructiontuning, wang2024qwen2vlenhancingvisionlanguagemodels}, but they are often limited to processing visual cues, neglecting the rich auditory information present in videos.
To address this, Omni-Modal Large Language Models (Omni-LLMs) \citep{shu2023audiovisualllmvideounderstanding, tang2025videosalmonn2captionenhancedaudiovisual, xu2025qwen25omnitechnicalreport, xu2025qwen3omnitechnicalreport} were developed to integrate visual, auditory, and textual modalities within a unified autoregressive architecture. By capturing intricate cross-modal relationships and contextual nuances, this paradigm significantly enhances the model's capacity to perceive and interpret complex environments.

\begin{wrapfigure}{r}{0.65\textwidth}
    \centering
    \vspace{-10pt}
     \includegraphics[width=0.65\textwidth,trim=50 27 45 25, clip]{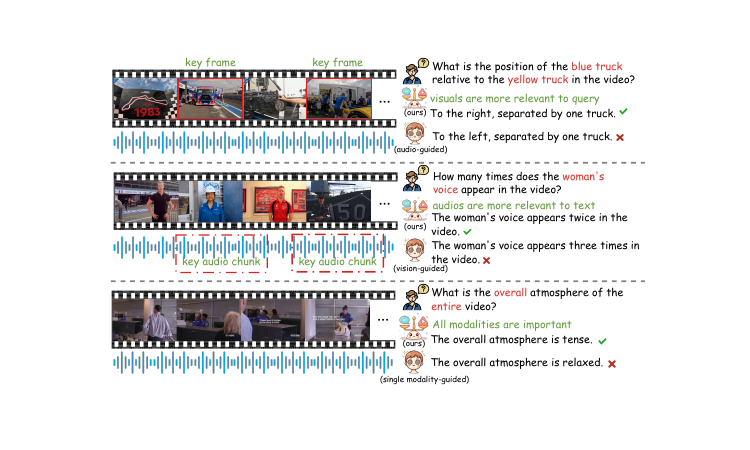}
    \vspace{-10pt}
    \caption{Illustration of modality importance variation across different questions.}
    \label{fig:intro_case}
    \vspace{-15pt}
\end{wrapfigure}

Since Omni-LLMs must process high-fidelity video and audio streams during inference, the resulting tokenization leads to an excessive number of tokens. This substantially increases the quadratic complexity of attention in Omni-LLMs, leading to significant computational and memory bottlenecks. To alleviate these issues, token compression has emerged as a promising technique. Substantial progress has already been made in single-modality compression involving images \citep{yang2026visionziplongerbetternecessary, ye2024fitprunefasttrainingfree}, videos \citep{shen2024longvuspatiotemporaladaptivecompression, tao2025dycokedynamiccompressiontokens}, and audios \citep{li2023acceleratingtransducersadjacenttoken, lin2025speechprunecontextawaretokenpruning}, which demonstrates its effectiveness in improving efficiency.

Nevertheless, very few works have addressed token compression within the context of Omni-LLMs. 
OmniZip \citep{tao2025omnizipaudioguideddynamictoken} introduces an innovative audio-visual token compression framework that utilizes audio anchors to dynamically guide video pruning through an interleaved spatiotemporal scheme, effectively accelerating OmniLLM inference in a training-free manner.
OmniSIFT \citep{ding2026omnisiftmodalityasymmetrictokencompression} introduces a modality-asymmetric token compression framework that employs a two-stage strategy of spatio-temporal video pruning and vision-guided audio selection, optimized end-to-end via a differentiable straight-through estimator. 

While OmniZip \citep{tao2025omnizipaudioguideddynamictoken} and OmniSIFT \citep{ding2026omnisiftmodalityasymmetrictokencompression}  utilize static guidance mechanisms based on audio or vision respectively, our analysis reveals that such one-size-fits-all strategies are inadequate. 
As shown in Figure~\ref{fig:intro_case}, the relative importance of visual and auditory information varies significantly across different questions. While some questions rely heavily on vision, others are driven by audio or require a balanced integration of both modalities. Employing an inappropriate modality to guide token compression can lead to erroneous model responses. Therefore, we argue that the guidance strategy for token compression should be selected dynamically rather than relying on a fixed modality.

To facilitate dynamic selection of the guidance modality for token compression, it is essential to identify which modality is more relevant to the question.
We employ AudioCLIP~\cite{guzhov2022audioclip} to evaluate the cross-modal correlations between the question and various modalities. 
In the field of video keyframe sampling, existing works such as Q-Frame \citep{zhang2025qframequeryawareframeselection} and AKS \citep{tang2025adaptivekeyframesamplinglong} also utilize CLIP-based vision-language models to compute the relative importance between questions and visual information. 
Based on these principles, we present \textbf{OmniSelect}, a training-free framework that utilizes a two-stage strategy to dynamically compress visual and audio tokens. First, we utilize AudioCLIP ~\cite{guzhov2022audioclip} to calculate the similarity scores of visual-text and audio-text pairs, which defines the pruning ratio and categorizes the strategy into three distinct types: Uniform Pruning, Video-Centric Pruning, and Audio-Centric Pruning. Second, redundant multimodal tokens are then pruned based on the attention score and cosine similarity matrix within each temporal group. 

Experimental results demonstrate that OmniSelect delivers exceptional performance on audio-visual understanding tasks, and achieves competitive performance among existing training-free token compression methods. As shown in Figure ~\ref{fig:overview}, OmniSelect achieves an inference speedup of 1.19$\times$ to 1.33$\times$ while reducing GPU memory consumption by 2.61GB to 2.81GB. Despite these substantial resource savings, OmniSelect retains 94\% to 99\% of the original full-token accuracy.

Overall, our contributions are as follows: 
\begin{itemize}
\item We analyze that existing methods typically rely on fixed, modality-specific guidance, which fails to account for the varying importance of modalities across different queries. Consequently, we propose the insight that modality-guided pruning should be selected dynamically.
\item We propose \textbf{OmniSelect}, a training-free, modality-adaptive token compression framework that dynamically selects appropriate compression strategies for multimodal inputs and performs fine-grained token pruning within each temporal group, adaptively allocating pruning ratios to preserve informative tokens across modalities.
\item Experimental results on audio-visual benchmarks show that OmniSelect accelerates inference speed and reduces GPU memory overhead while maintaining high performance and achieving competitive performance among  training-free token compression methods.
\end{itemize}

\section{Related Work}

\subsection{Omni-modal Large Language Models}

Omni-Modal Large Language Models (Omni-LLMs) \citep{jiang2025specificmllmsomnimllmssurveymllms} represent an advanced evolution of traditional VideoLLMs \citep{an2025llavaonevision15fullyopenframework, bai2025qwen25vltechnicalreport}, incorporating audio alongside visual and textual data within a unified autoregressive architecture. This comprehensive approach enables the models to better understand complex environments by capturing intricate inter-modal dependencies and contextual nuances \citep{shu2023audiovisualllmvideounderstanding, cheng2024videollama2advancingspatialtemporal, yang2025humanomniv2understandingomnimodalreasoning, li2025baichuanomni15technicalreport, tang2025videosalmonn2captionenhancedaudiovisual, tong2025interactiveomniunifiedomnimodalmodel}.
Leading proprietary systems such as Qwen 3.5-Omni \citep{qwen35omniblog} and GPT-4o \citep{openai2024gpt4ocard} have set high standards in audio-visual comprehension benchmarks \citep{zhou2026dailyomniaudiovisualreasoningtemporal, hong2026worldsenseevaluatingrealworldomnimodal}. Simultaneously, the open-source community has created notable models such as Qwen2.5-Omni \citep{xu2025qwen25omnitechnicalreport} and Qwen3-Omni \citep{xu2025qwen3omnitechnicalreport}. These models use an end-to-end perception strategy that aligns specialized modality encoders with a central large language model (LLM) backbone through optimized projection layers.

\begin{wrapfigure}{r}{0.60\textwidth}
    \centering
    \vspace{-10pt}
    \includegraphics[width=0.60\textwidth,trim=64 45 105 43, clip]{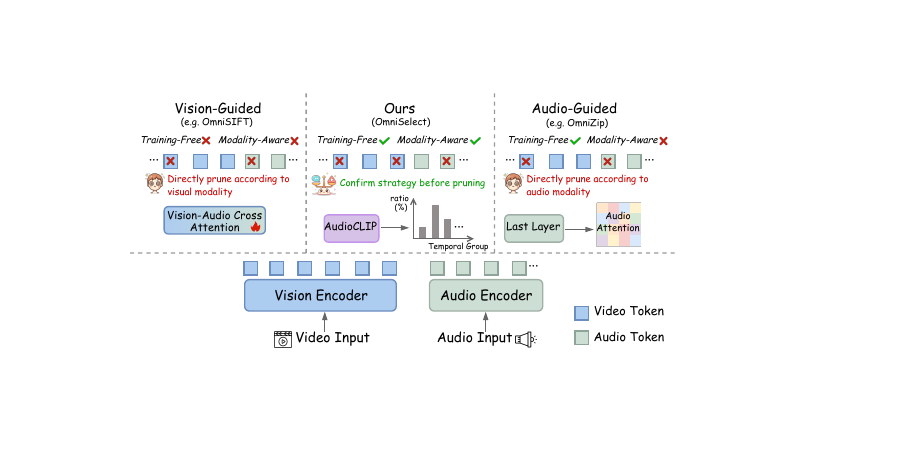}
    \vspace{-10pt}
    \caption{Illustration of different token compression strategies: vision-guided, audio-guided, and ours.}
    \label{fig:compress_compare}
    \vspace{-15pt}
\end{wrapfigure}

\subsection{Token Compression in Omni-LLMs}

Processing high-fidelity video and audio streams generates an overwhelming number of multimodal tokens, hindering efficient deployment. The main challenge now is to maintain high-level reasoning performance while creating techniques to compress or simplify these integrated audio-visual inputs to reduce computational overhead.
While token compression methods for images \citep{yang2026visionziplongerbetternecessary, ye2024fitprunefasttrainingfree}, videos \citep{shen2024longvuspatiotemporaladaptivecompression, tao2025dycokedynamiccompressiontokens}, and audios \citep{li2023acceleratingtransducersadjacenttoken, lin2025speechprunecontextawaretokenpruning} tasks have been widely studied, recent research has paid little attention to token compression in the omnimodal setting.
As shown in Figure~\ref{fig:compress_compare},
Omnizip \citep{tao2025omnizipaudioguideddynamictoken} identifies salient audio tokens and computes an audio retention score for each time group, and uses them to guide video token pruning.
OmniSIFT \citep{ding2026omnisiftmodalityasymmetrictokencompression} introduces a two-stage compression framework with spatio-temporal video pruning and vision-guided audio selection, optimized end-to-end via a differentiable straight-through estimator. 
Existing methods typically rely on fixed, modality-specific guidance, which fails to account for the varying importance of modalities across different queries. To address this limitation, our approach is based on Modality-Aware techniques that dynamically determine the best pruning strategies.

\section{Methodology}

\subsection{Our Method: OmniSelect}
\begin{figure*}[t]
    \centering
    \includegraphics[width=1.0\textwidth,trim=100 158 100 158, clip]{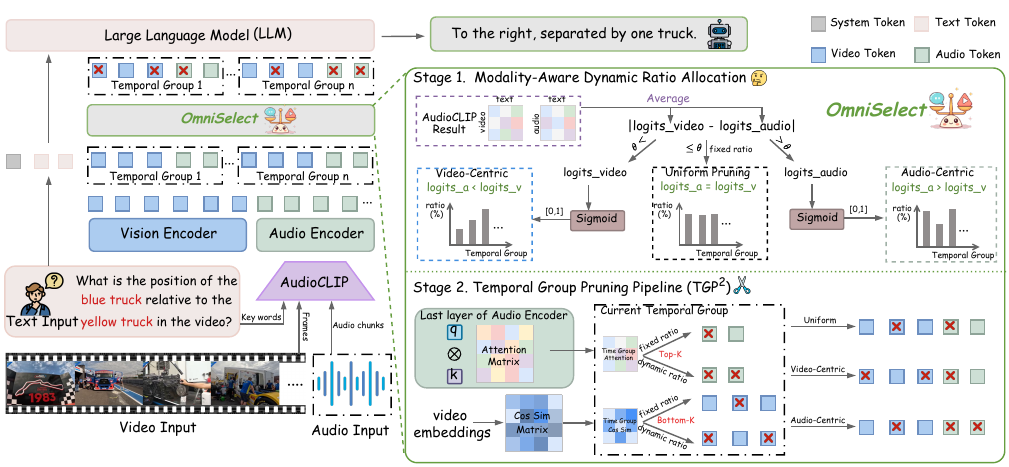}
    \caption{Overview pipeline of our method \textbf{OmniSelect}. The overall process is divided into two stages: (1) Modality-Aware Dynamic Ratio Allocation that allocates the pruning ratio for each temporal group while ensuring the total pruning ratio meets the expected value; (2) \textbf{T}emporal \textbf{G}roup \textbf{P}runing \textbf{P}ipeline (TGP$^{2}$) that prunes audio and video tokens based on attention score and cosine similarity score within each temporal group.}
    \label{fig:method}
\end{figure*}

As illustrated in Figure~\ref{fig:method}, our method \textbf{OmniSelect} is fully training-free and utilizes a two-stage strategy to dynamically compress visual and audio tokens. First, we calculate the similarity scores of visual-text and audio-text pairs, map the resulting logits to the range \([0, 1]\), and use them to define the pruning ratio \(\rho_i\) for each temporal group \(i\). This step also categorizes the pruning strategy into three distinct types: Uniform Pruning, Video-Centric Pruning, and Audio-Centric Pruning. Second, the method employs the \textbf{T}emporal \textbf{G}roup \textbf{P}runing \textbf{P}ipeline (\textbf{TGP}$^{2}$). After determining the pruning strategy, redundant multimodal tokens are pruned based on attention scores and cosine similarity matrix within each temporal group \(i\). Notably, we set fixed pruning ratios \(\rho_v\) and \(\rho_a\) for visual and audio tokens, resulting in retained ratios of \(1 - \rho_v\) and \(1 - \rho_a\), respectively.

\subsection{Modality-Aware Dynamic Ratio Allocation}
\label{sec: modalityaware}

Previous methods usually perform token pruning based on a single modality. However, in omni-modal reasoning, different modalities often contribute unequally depending on the query. Motivated by this observation, we aim to dynamically determine whether visual or audio information is more informative or whether both modalities should be treated equally. Preliminary analysis is provided in Appendix~\ref{sec: preliminary}. As discussed in this section, visual and audio tokens are grouped into temporal chunks before entering the LLM backbone. Following this design, we allocate pruning ratios at the chunk level, where each audio segment is temporally aligned with a sampled video frame. 

To estimate modality importance, we compute cross-modal similarities between visual, audio, and textual representations using AudioCLIP~\cite{guzhov2022audioclip}, which maps all modalities into a shared embedding space. To reduce computational cost, video frames are temporally downsampled. For text, we retain only nouns and adjectives through lightweight linguistic filtering, since these words mainly capture semantic objects and attributes. The similarity scores are computed as:
\begin{flalign}
& s_i^{(v)} = \frac{\mathbf{v}_i \cdot \mathbf{t}}{\|\mathbf{v}_i\| \, \|\mathbf{t}\|},
\quad
s_i^{(a)} = \frac{\mathbf{a}_i \cdot \mathbf{t}}{\|\mathbf{a}_i\| \, \|\mathbf{t}\|}, \notag\\
&\bar{s}^{(v)} = \frac{1}{F} \sum_{i=1}^{F} s_i^{(v)},
\quad
\bar{s}^{(a)} = \frac{1}{F} \sum_{i=1}^{F} s_i^{(a)}, \quad i = 1, 2, \dots, F,
\end{flalign}
where $\{\mathbf{v}_i\}_{i=1}^{F}$, $\{\mathbf{a}_i\}_{i=1}^{F}$, and $\mathbf{t}$ denote the visual embeddings of the $F$ video frames, audio embeddings of the $F$ audio chunks, and textual embeddings, respectively. The averaged scores $\bar{s}^{(v)}$ and $\bar{s}^{(a)}$ reflect the overall relevance of the visual and audio modalities to the query, respectively.

Based on the score difference and a threshold \(\theta\), we divide the pruning process into three cases. If \(|\bar{s}^{(v)} - \bar{s}^{(a)}| \leq \theta\), both modalities are considered equally important and we apply \textbf{Uniform Pruning}. Otherwise, pruning becomes modality-aware. When \(\bar{s}^{(v)} > \bar{s}^{(a)}\), we adopt \textbf{Video-Centric Pruning}; otherwise, we use \textbf{Audio-Centric Pruning}. For the less informative modality, we still employ Uniform Pruning to avoid excessive information loss.

To satisfy the global pruning budget while preserving score-based importance, we further design an adaptive allocation strategy. Let \(\mathbf{s} = \{\hat{s_1}, \hat{s_2}, \dots, \hat{s_G}\}\) denote the similarity scores and \(\mathbf{n} = \{n_1, n_2, \dots, n_G\}\) the token counts of the \(G\) temporal groups. The expected number of pruned tokens is defined as \(N_{exp} =  \rho_{v/a} \cdot \sum_{i=1}^{G} n_i\), where  \(\rho_{v/a}\) is the target pruning ratio for audio tokens or video tokens.

We first normalize the scores using their mean \(\mu_s\) and standard deviation \(\sigma_s\). When \(\sigma_s > \epsilon\), the base pruning probability is obtained through a sigmoid mapping:
\begin{flalign}
    & p_i = \sigma \left( -\frac{\hat s_i - \mu_s}{\tau \cdot \sigma_s} \right), \notag\\
    & \sigma(x) = \frac{1}{1 + \exp(-x)},
\end{flalign}
where \(\tau\) controls the sharpness of the pruning distribution. If \(\sigma_s \le \epsilon\), all groups share the same pruning probability \(p_i = \eta\). Next, we rescale the probabilities to match the global pruning budget and iteratively refine the allocation:
\begin{flalign}
    &\rho_i^{(0)} = \text{clip} \left( p_i \cdot \frac{N_{exp}}{\sum_{j=1}^{G} p_j n_j}, 0, 1 \right), \notag\\
    &\rho_i^{(k+1)} = \text{clip} \left( \rho_i^{(k)} + \frac{N_{exp} - \sum_{j=1}^{G} \rho_j^{(k)} n_j}{\sum_{j \in \mathcal{A}} n_j}, 0, 1 \right), \quad \forall i \in \mathcal{A},
\end{flalign}
where \(\mathcal{A} = \{i \mid 0 < \rho_i^{(k)} < 1\}\). The refinement continues until the pruning budget is satisfied or the maximum iteration number is reached. Finally, the pruning ratio for the \(i\)-th temporal group in modality \(m \in \{v, a\}\) is defined as:
\begin{equation}
\rho_{i, m} =
\begin{cases}
\rho_{v/a}, & |\bar{s}^{(v)} - \bar{s}^{(a)}| \leq \theta, \\
\sigma \left( -\frac{\hat s_{i,v} - \mu_{s_{v}}}{\tau \cdot \sigma_{s_{v}}} \right), & |\bar{s}^{(v)} - \bar{s}^{(a)}| > \theta \quad \text{and} \quad \bar{s}^{(v)} > \bar{s}^{(a)},\\
\sigma\left( -\frac{\hat s_{i, a} - \mu_{s_{a}}}{\tau \cdot \sigma_{s_{a}}} \right), & |\bar{s}^{(a)} - \bar{s}^{(v)}| > \theta \quad \text{and} \quad \bar{s}^{(v)} < \bar{s}^{(a)},
\end{cases}
\end{equation}
where \(\tau\) is a temperature hyperparameter controlling the concentration of pruning ratios. The retained token number for each temporal group is then computed as:
\begin{equation}
K_{i,m} = \max \left( 1, \lfloor (1 - \rho_{i,m}\%) \cdot n_i \rfloor \right).
\end{equation}

Overall, this strategy dynamically allocates computational resources toward the modality that is more semantically aligned with the query while maintaining a global pruning constraint.

\subsection{Temporal Group Pruning Pipeline (TGP$^{2}$)}
After allocating the pruning ratios for each temporal group, we prune the visual and audio tokens within each temporal group to retain more salient tokens before model inference. To achieve this, we propose the \textbf{T}emporal \textbf{G}roup \textbf{P}runing \textbf{P}ipeline (\textbf{TGP}$^{2}$). For the audio token pruning strategy, we compute the attention matrix from the last layer of the audio encoder \(\mathcal{E}_{a}\) as follows:
\begin{flalign}
    & \mathbf{Q}_a = \mathbf{W}_{\mathbf{Q}_{a}}(\mathcal{T}_{a}), \quad \mathbf{K}_a = \mathbf{W}_{\mathbf{K}_{a}}(\mathcal{T}_{a}), \notag\\
    & \mathbf{A}^{(a)} = \text{Softmax} \left( \frac{\mathbf{Q}_a \mathbf{K}_a^{\top}}{\sqrt{d_k}} \right) \in \mathbb{R}^{n_{i,a} \times n_{i,a}},
\end{flalign}
where \(n_{i,a}\) represents the total number of audio tokens in the temporal group \(i\). After performing the same pooling operation as in the encoder to align with actual audio token indices, the averaged attention score is denoted as \(\mathbf{A}^{(a)}_{avg}\). Finally, we select the salient audio tokens by:
\begin{flalign}
    & \mathcal{I}_{i,a} = \text{TopK} \left(\mathbf{A}^{(a)}_{avg}, \ k = \lfloor (1-\rho_{i,a}\%) \cdot (\lceil \frac{n_{i,a}}{2}\rceil)\rfloor \right).
\end{flalign}

For the vision token pruning strategy, we compute the cosine similarity matrix of the vision embeddings \(\mathbf{V}_i\) in the \(i\)-th temporal group as \(\mathbf{S} = \bar{\mathbf{V}}_i \bar{\mathbf{V}}_i^\top\), where \(\bar{\mathbf{V}}_i\) denotes the \(\ell_2\)-normalized embeddings. We then compute the average similarity score for each token by averaging across the last dimension. The indices of salient tokens \(\mathcal{I}_{i,v}\) are selected via:
\begin{flalign}
    \mathcal{I}_{i,v} = \text{BottomK} \left( \frac{1}{N_v} \sum_{j=1}^{N_v} \mathbf{S}_{v,j}, \quad k = \lfloor (1-\rho_{i,v}\%) \cdot n_{i,v} \rfloor \right).
\end{flalign}

 We use video embeddings instead of attention weights because they are more computationally efficient and provide a more reliable measure of token redundancy without the attention sink bias. More detailed analyses about why to choose the Bottom-K strategy are provided in the Appendix~\ref{more imp details}.

\section{Experiments}

\begin{table*}[t]
\centering
\captionsetup{skip=4pt}
\renewcommand{\arraystretch}{0.94} %
\setlength{\tabcolsep}{4pt} %
\caption{
   \textbf{Main results on OmniVideoBench, VideoMME, and DailyOmni.} \textbf{*} denotes performance exceeding Full Tokens. The best and second-best results are \textbf{bolded} and \textbf{underlined} for each column, respectively.
}
\resizebox{1\linewidth}{!}{
\begin{tabular}{cccccccccccccc}

\toprule
\multirow{2}{*}{Method} 
& \multirow{2}{*}{\makecell{Retained\\Ratio (\%)}} 
& \multirow{2}{*}{\makecell{OmniVideo\\Bench ($\uparrow$)}} 
& \multicolumn{4}{c}{VideoMME ($\uparrow$)} 
& \multicolumn{7}{c}{DailyOmni ($\uparrow$)} \\
\cmidrule(lr){4-7} \cmidrule(lr){8-14}
& & & Short & Medium & Long & Avg. & Con. & Event & AV Event & Com. & Inf. & Rea. & Avg.\\ 
\midrule
\multicolumn{14}{c}{\emph{Qwen2.5-Omni-3B}} 
\\ 
\midrule
\rowcolor{mycol3}
\multicolumn{14}{l}{\textit{Frame Budgets 128 (VideoMME 512):}} \\
\rowcolor{mycol2!40!white}
Full Tokens & 100\% & 32.9 & 74.8 & 64.1 & 52.8 & 63.89 & 55.44 & 56.21 &  53.78 & 70.99 & 79.22 & 74.29 & 62.82 \\
Random & 55\% & 31.0 & 68.8 & 61.6 & 52.4 & 61.00 & 47.67 &  44.44 &  45.38 & 64.12 & 71.43 & 63.43 & 53.55\\
DyCoke (V\&A) & 50\% & 31.3 &  71.1 & 62.8 & 52.4 & 62.11 & \underline{48.19} &  \underline{49.35} &  47.48 &  \textbf{65.65} & 72.73 & 65.14 & 55.89\\
OmniZip & 45\% & 32.5 & \textbf{74.0} & \underline{64.6} & 52.7 & \underline{63.74} & 47.15 &  47.71 &  46.64 & \underline{64.12} & \textbf{76.62} & \textbf{69.71} & \underline{56.14}\\
OmniZip & 30\%  & 31.4 & 70.6 & 61.2 & 52.7 & 61.48 & 46.63 &  46.08 &  42.86 &  59.54 & 68.18 & 68.00 & 53.05\\
\rowcolor{mycol1}
OmniSelect (Ours) & 45\% & \underline{32.7} & \underline{73.4} & \textbf{65.0}$^{*}$ & \textbf{53.3}$^{*}$ &  \textbf{63.93}$^{*}$ & \textbf{50.78} &  \textbf{49.67} & \textbf{52.10} & \textbf{65.65} & \underline{74.03} & \underline{69.14} & \textbf{58.06}\\ 
\rowcolor{mycol1}
OmniSelect (Ours) & 30\% & \textbf{33.3}$^{*}$ & 69.7 & 61.9 &  \underline{53.2} &  61.59 & 46.63 &  45.42 &  \underline{48.32} & 62.60 & 72.08 & 65.14 & 54.39\\
\midrule
\multicolumn{14}{c}{\emph{Qwen2.5-Omni-7B}} 
\\ 
\midrule
\rowcolor{mycol3}
\multicolumn{14}{l}{\textit{Frame Budgets 128 (VideoMME 512):}} \\
\rowcolor{mycol2!40!white}
Full Tokens & 100\% & 34.6 & 77.1 & 66.8 & 55.4 & 66.44 & 56.99 & 60.13 &  50.84 &  71.76 & 78.57 & 78.86 & 64.16\\
Random & 55\% & 32.1 & 73.9 & 66.6 & 55.3 & 65.07 & 52.33 &  49.67 &  41.48 &  \underline{70.99} & 72.73 & 71.43 & 56.89\\
DyCoke (V\&A) & 50\% & 32.4 & 73.9 & 66.4 & 55.1 & 65.15 & 50.26 &  \underline{52.94} &  44.96 &  70.23 & 77.92 & \textbf{76.57} & 59.48\\
OmniZip & 45\% & \textbf{33.3} & \textbf{75.4} & \underline{67.2} & 55.4 & \underline{66.03} & \textbf{53.89} & 51.96 &  \underline{46.22} & 70.23 & \textbf{77.27} & \textbf{76.57} & \underline{59.98}\\
OmniZip & 30\%  &  32.8 & 74.3 & 66.1 & 55.4 & 65.29 & 47.15 &  47.39 &  43.70 & 63.36 & \underline{76.62} & \underline{73.71} & 55.97\\
\rowcolor{mycol1}
OmniSelect (Ours) & 45\% & \underline{33.1} & \underline{74.9} &  \textbf{67.7}$^{*}$ & \textbf{56.4}$^{*}$ & \textbf{66.33} & \underline{53.37} & \textbf{53.92} &  \textbf{46.64} & \textbf{72.52}$^{*}$ & \underline{76.62} & \textbf{76.57} & \textbf{60.65}\\
\rowcolor{mycol1}
OmniSelect (Ours) & 30\%  & 32.7 & 73.8 & 66.3 & \underline{56.3} & 65.48 &  48.19 & 50.00 &  \underline{46.22} & 67.18 & 74.68 & 73.14 & 57.39\\
\bottomrule
\end{tabular}
}
\label{tab:mainresult}
\vspace{-3mm}
\end{table*}

\begin{table*}[t]
\centering
\captionsetup{skip=4pt}
\renewcommand{\arraystretch}{0.94} %
\setlength{\tabcolsep}{4pt} %
\caption{{
\textbf{WorldSense results.} \textbf{*} denotes performance exceeding Full Tokens. The best and second-best results are \textbf{bolded} and \textbf{underlined} for each column, respectively.
}
}
\resizebox{1\linewidth}{!}{
\begin{tabular}{ccccccccccc}
\toprule
Method & Retained Ratio & \begin{tabular}[c]{@{}c@{}}Tech \&\\ Science\end{tabular} & Games & \begin{tabular}[c]{@{}c@{}}Daily\\ Life\end{tabular} & \begin{tabular}[c]{@{}c@{}}Film \&\\ TV\end{tabular} & Music & Sports & \begin{tabular}[c]{@{}c@{}}Culture \&\\ Politics\end{tabular} & Performance & Avg. ($\uparrow$) \\ \midrule
\multicolumn{11}{c}{\emph{Qwen2.5-Omni-3B}} 
\\ 
\midrule
\rowcolor{mycol3}
\multicolumn{11}{l}{\textit{Frame Budgets 32:}} \\
\rowcolor{mycol2!50!white}
Full Tokens & 100\% & 49.39 & 37.77 & 44.07 & 42.48 & 45.32 & 39.77 & 47.57 & 38.20 & 43.66 \\
OmniZip & 45\%  &  \textbf{47.35} & 37.77 & \underline{41.64} & 39.58 & 43.10 & \underline{37.67} & 42.39 &  36.33 & 41.27 \\
OmniZip & 30\%  &  46.12 & \textbf{39.48}$^{*}$ & 39.51 & 36.15 & \underline{44.58} & 35.35 & 38.83 &  34.83 & 39.75 \\
\rowcolor{mycol1}
OmniSelect (Ours) & 45\% & \underline{46.94} & \underline{38.20} & \textbf{42.86} &  \textbf{40.90} &  44.33 & \textbf{39.30} & \underline{44.98} & \underline{37.45} &  \textbf{42.37} \\
\rowcolor{mycol1}
OmniSelect (Ours) & 30\% & \underline{46.94} & 36.48 & 41.19 & \underline{40.63} &  \textbf{47.04}$^{*}$ & 36.74 &  \textbf{45.95} & \textbf{37.83} & \underline{41.99} \\ 
\rowcolor{mycol3}
\multicolumn{11}{l}{\textit{Frame Budgets 64:}} \\
\rowcolor{mycol2!50!white}
Full Tokens & 100\%  & 51.02 & 39.48 & 43.62 & 42.22 & 45.32 & 41.63 & 50.16 & 43.45 & 44.86 \\
OmniZip & 45\% &  \textbf{48.78} & 38.63 & \underline{43.16} & \underline{40.11} & \textbf{45.32} & \underline{38.84} & 44.66 &  \textbf{39.33} & \underline{42.84} \\
OmniZip & 30\%  &  \underline{46.33} & \underline{39.06} & 40.12 & 37.47 & 44.33 & 36.05 & 42.07 &  37.08 & 40.61 \\
\rowcolor{mycol1}
OmniSelect (Ours) & 45\% & \textbf{48.78} & {38.20} & \textbf{43.62}$^{*}$ & \textbf{41.69} &  \underline{44.58} & \textbf{39.53} &  \textbf{46.93} & \underline{37.83} & \textbf{43.19} \\ 
\rowcolor{mycol1}
OmniSelect (Ours) & 30\% & \textbf{48.78} & \textbf{41.63}$^{*}$ & {41.64} &  39.84 & 44.33 & 38.37 & \underline{46.28} & \underline{37.83} &  42.56 \\
\rowcolor{mycol3}
\multicolumn{11}{l}{\textit{Frame Budgets 128:}} \\
\rowcolor{mycol2!40!white}
Full Tokens & 100\% & 52.65 & 39.06 & 43.47 & 44.06 & 46.55 & 42.79 & 52.43 & 41.20 & 45.62 \\
Random & 55\% & 46.53 & 36.05 & 40.27 & 38.79 & 45.81 & 39.53 & 46.60 & 36.70 & 41.68 \\
DyCoke (V\&A) & 50\% & 49.18 &  \underline{40.34} & 42.10 & 39.84 & 43.60 & \underline{41.86} & 45.95 & 38.95 & 43.06 \\
OmniZip & 45\% &  \underline{50.20} & 39.91 & 43.16 & 41.16 & \underline{46.55} & 40.23 & \underline{48.54} &  40.07 & 44.07 \\
OmniZip & 30\%  &  47.76 & \textbf{41.20}$^{*}$ & 40.88 & 38.52 & {45.57} & 37.44 & 44.01 &  37.45 & 41.83 \\
\rowcolor{mycol1}
OmniSelect (Ours) & 45\% & \textbf{50.61} & \underline{40.34} & \underline{43.77} & \textbf{43.27} &  46.31 & \textbf{42.56} &  \textbf{49.51} & \textbf{41.95}$^{*}$ & \textbf{45.08} \\ 
\rowcolor{mycol1}
OmniSelect (Ours) & 30\% & 49.80 & \textbf{41.20}$^{*}$ & \textbf{43.92}$^{*}$ &  \underline{43.01} &  \textbf{46.80}$^{*}$ & 39.53 & {47.57} & \underline{41.20} &  \underline{44.42} \\
\midrule
\multicolumn{11}{c}{\emph{Qwen2.5-Omni-7B}} 
\\ 
\midrule
\rowcolor{mycol3}
\multicolumn{11}{l}{\textit{Frame Budgets 128:}} \\
\rowcolor{mycol2!40!white}
Full Tokens & 100\% & 49.39 & 42.49 & 46.96 & 44.59 & 48.03 & 42.56 & 53.72 & 45.69 & 46.82 \\
Random & 55\% & \underline{47.76} & 36.48 & 42.86 & 40.90 & \underline{48.28} & 38.14 & 49.19 & {38.95} & 43.25 \\
DyCoke (V\&A) & 50\% & 46.73 & 40.77 & 43.62 & 39.58 & 47.29 & {41.86} & 48.54 & 41.57 & 43.95 \\
OmniZip & 45\% & 47.35 & \underline{41.20} & \underline{45.14} & 42.74 & \textbf{49.01}$^{*}$ & 42.33 & \underline{49.51} & \underline{43.07} &  \underline{45.27} \\
OmniZip & 30\%  &  45.92 & 39.91 & 43.92 & \textbf{43.27} & 48.03 & 40.23 & 46.93 &  40.07 & 43.85 \\
\rowcolor{mycol1}
OmniSelect (Ours) & 45\% & \textbf{47.96} & \textbf{43.78}$^{*}$ & \textbf{46.20} &  \underline{43.01} & 48.03 & \textbf{43.72}$^{*}$ & \textbf{50.81} &  \underline{43.07} & \textbf{46.00} \\
\rowcolor{mycol1}
OmniSelect (Ours) & 30\%  & 46.33 & 39.91 & 44.38 & {42.22} & 45.32 &  \underline{42.79} &  46.93 &  \textbf{43.45} & 44.17 \\
\bottomrule
\end{tabular}
}
\label{tab:WorldSense}
\vspace{-3mm}
\end{table*}

\subsection{Experiments Setup}
\label{sec:4.1}
\textbf{Models and Baselines.} 
We conduct experiments on Qwen2.5-Omni models with 3B and 7B scales. We compare our method OmniSelect with three baselines. (i) \textbf{OmniZip}~\citep{tao2025omnizipaudioguideddynamictoken} is a training-free framework that dynamically compresses multimodal tokens by leveraging salient audio cues to guide video token pruning. (ii) \textbf{Random Pruning} randomly removes audio and video tokens under the same pruning ratio. We include this baseline to verify the effectiveness of our structured pruning strategy. (iii) \textbf{DyCoke}~\citep{tao2025dycoke} is a training-free dynamic token pruning method for video LLMs. In our experiments, we apply its first-stage Token Temporal Merging (TTM) module to compress both video and audio tokens. We do not include OmniSIFT~\citep{tao2025omnizipaudioguideddynamictoken} in the primary comparison because it requires training with a straight-through estimator, whereas our comparison focuses on training-free token compression methods.

\textbf{Benchmarks.} 
We evaluate OmniSelect and the baselines on four benchmarks: WorldSense~\citep{hong2026worldsenseevaluatingrealworldomnimodal}, DailyOmni~\citep{zhou2026dailyomniaudiovisualreasoningtemporal}, VideoMME (with audio)~\citep{fu2025video}, and OmniVideoBench~\citep{li2025omnivideobench}. WorldSense contains 3,172 QA pairs and focuses on real-world omnimodal understanding, requiring joint reasoning over audio, visual, and textual signals across diverse scenarios. DailyOmni consists of 1,197 samples and emphasizes audio-visual reasoning in daily-life scenarios with strong temporal dependencies. OmniVideoBench includes 1,000 QA pairs and is designed to evaluate synergistic audio-visual reasoning with step-by-step annotations, highlighting modality complementarity and logical consistency. VideoMME comprises approximately 2,700 samples and provides a comprehensive evaluation of video understanding, covering perception, temporal reasoning, and multimodal integration.

\textbf{Implementation Details.} 
When applying OmniSelect, we set the threshold \(\theta\) following Section~\ref{sec:4.4}. Specifically, we use \(\theta = 0\) when \(|\bar{s}^{(v)} - \bar{s}^{(a)}| \leq 2\), and \(\theta = 5\) when \(|\bar{s}^{(v)} - \bar{s}^{(a)}| > 2\). This setting is used across all benchmarks in Tables~\ref{tab:mainresult} and~\ref{tab:WorldSense}, providing stable performance across datasets. Note that a single threshold is not optimal for all cases due to varying modality dominance, and per-instance best results are reported in Appendix~\ref{best results}. We evaluate robustness under different frame budgets on WorldSense and DailyOmni, while setting 512 frames for VideoMME and 128 frames for OmniVideoBench. FlashAttention is used in all experiments to improve inference efficiency and reduce memory overhead. Additional implementation details are provided in Appendix~\ref{more imp details}.

\begin{figure}[t]
    \centering
    \begin{subfigure}[b]{0.49\textwidth}
        \centering
        \includegraphics[width=1.0\textwidth,trim=5 6 5 5, clip]{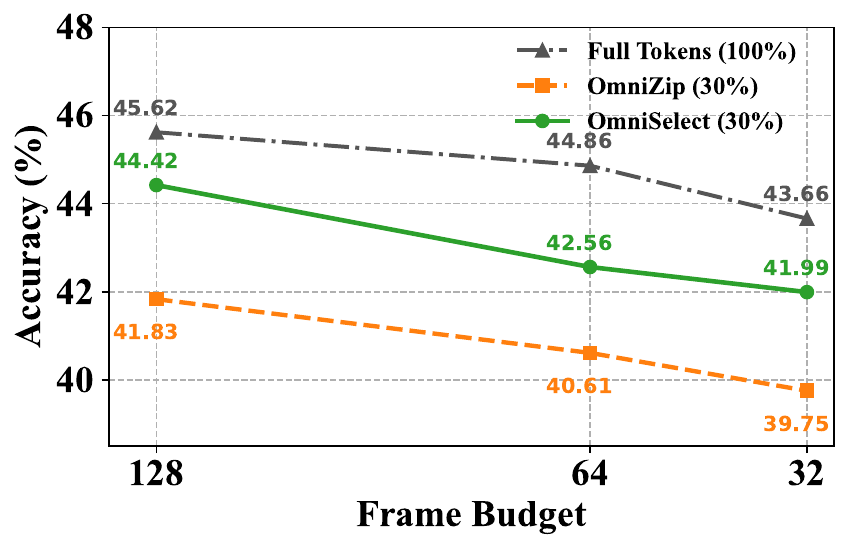}
        \label{fig:budget_30}
    \end{subfigure}
    \hfill
    \begin{subfigure}[b]{0.49\textwidth}
        \centering
        \includegraphics[width=\textwidth,trim=5 6 5 5, clip]{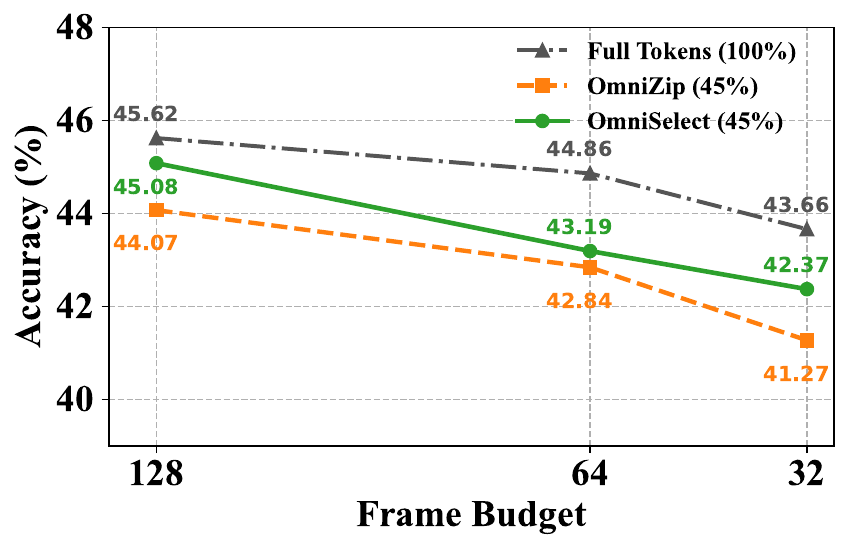}  
        \label{fig:budget_45}
    \end{subfigure}
    
    \caption{Qwen-2.5-Omni-3B performance under varying frame budgets at 30\% and 45\% pruning.}
    \label{fig:budgets}
\end{figure}

\subsection{Main Results}

\textbf{Competitive Performance among Training-Free Compression Methods.} 
As shown in Table~\ref{tab:mainresult} and Table~\ref{tab:WorldSense}, when retaining 45\% of tokens, OmniSelect surpasses all training-free compression baselines on most datasets. It preserves 97.7\% of the full-token performance on OmniVideoBench across all compression settings and achieves 99.95\% performance retention on VideoMME at 45\% token retention. Compared with existing methods, OmniZip may underperform when visual information is more critical due to its audio-guided pruning strategy, while DyCoke can mistakenly compress salient tokens because of only temporal merging. In contrast, OmniSelect occasionally even surpasses full-token inference, suggesting that removing redundant tokens can improve prediction quality. Moreover, the advantage becomes more pronounced on long videos, where accurate selection of key frames and salient audio segments is particularly important.

\begin{figure}[t]
    \centering
    \begin{subfigure}[b]{0.49\textwidth}
        \centering
        \includegraphics[width=1.0\textwidth,trim=6 6 5 7, clip]{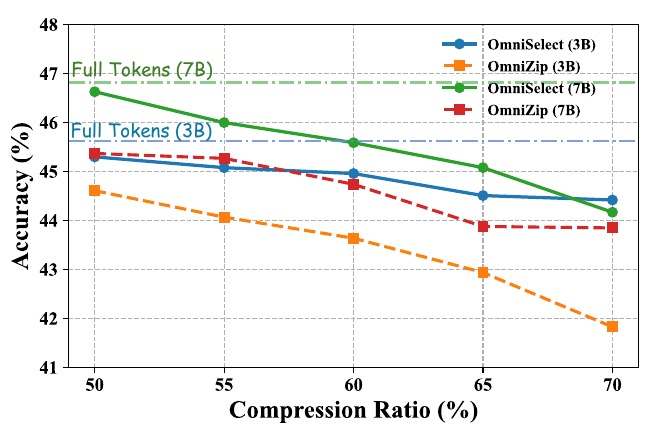}
        \caption{Performance under different compression ratios.}
        \label{fig:ratio}
    \end{subfigure}
    \hfill
    \begin{subfigure}[b]{0.49\textwidth}
        \centering
        \includegraphics[width=1.0\textwidth,trim=6 6 5 7, clip]{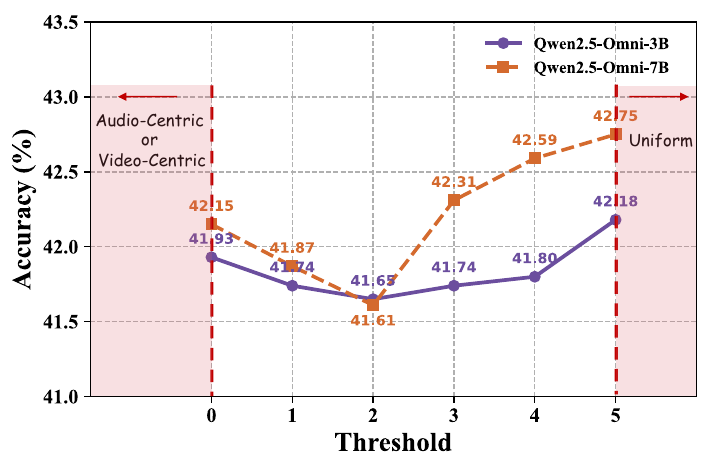} 
        \caption{Performance under different thresholds.}
        \label{fig:threshold}
    \end{subfigure}
    \caption{Performance under different compression ratios and different thresholds.}
\end{figure}

\textbf{Robustness across Frame Budgets and Compression Ratios.}
Existing omni-modal token compression methods usually adopt fixed frame budgets. To evaluate the robustness of OmniSelect, we conduct experiments on the WorldSense benchmark under varying frame budgets. As shown in Figure~\ref{fig:budgets}, at both 30\% and 45\% compression ratios, OmniSelect consistently outperforms OmniZip across 32, 64, and 128 frame budgets. This indicates that OmniSelect preserves informative tokens more effectively under limited input. We further evaluate performance under different compression ratios on WorldSense. As shown in Figure~\ref{fig:ratio}, OmniSelect consistently surpasses OmniZip across all tested ratios. Interestingly, we observe that token pruning yields larger gains on the 3B model compared to the 7B model, indicating a more favorable efficiency–performance trade-off in smaller backbones. This observation is consistent with the phenomenon reported in OmniZip~\citep{tao2025omnizipaudioguideddynamictoken}.

\subsection{Efficiency Analyses}

To further validate the effectiveness of OmniSelect, we evaluate the average peak GPU memory usage, average prefilling time, and average inference latency on Qwen2.5-Omni-3B and 7B models. Specifically, the latency measurements in Table\ref{tab:efficiency} use the same starting point for all methods after common raw multimodal data loading and preprocessing, while including all method-specific computation. Measured separately on WorldSense, AudioCLIP feature extraction and modality scoring require an average of 0.191 seconds per sample. After including this overhead, OmniSelect still reduces total latency. For OmniSelect, the reported latency already includes AudioCLIP feature extraction, similarity computation, strategy selection, and token pruning. As shown in Table~\ref{tab:efficiency}, OmniSelect achieves a GPU memory reduction of 2.58G–2.77G and provides a speedup of 1.19$\times$--1.33$\times$ during inference when compressing 70\% of the audio and video tokens. It outperforms both OmniZip and DyCoke while maintaining the highest accuracy among all methods. Regarding prefilling efficiency, OmniSelect substantially reduces the computational overhead during prompt prefilling, achieving 3.53$\times$-3.62$\times$ speedup under 70\% token compression. This gain mainly comes from early removal of redundant multimodal tokens, which reduces unnecessary self-attention computation in later transformer layers. 

\begin{wraptable}{r}{0.60\textwidth}
    \centering   
    \vspace{-15pt}
    \captionsetup{skip=4pt}
\renewcommand{\arraystretch}{0.94}
\setlength{\tabcolsep}{4pt}

\caption{\textbf{Efficiency Analyses on WorldSense benchmark}. The best result among the compression methods is \textbf{bolded}.}

\resizebox{1\linewidth}{!}{
\begin{tabular}{cccccc}
\toprule
Method & \makecell{Retained\\Ratio (\%)} & \makecell{GPU Mem.\\(G) $\downarrow$} & \makecell{Prefilling\\Time (ms) $\downarrow$} & \makecell{Average\\Latency (s) $\downarrow$} & Acc. (\%) $\uparrow$ \\
\midrule
\multicolumn{6}{c}{\emph{Qwen2.5-Omni-3B}} \\
\midrule
\rowcolor{mycol3}
\multicolumn{6}{l}{\textit{Frame Budgets 128:}} \\
\rowcolor{mycol2!40!white}
Full Tokens & 100\% & 15.08 & 594 ($1.00\times$) & 3.13 ($1.00\times$) & 45.62 \\
DyCoke (V\&A) & 50\% & 13.16 & 282 ($2.11\times$) & 2.93 ($1.07\times$) & 43.06 \\
OmniZip & 30\% & 12.49 & \textbf{163} ($3.64\times$) & 2.80 ($1.12\times$) & 41.83 \\
\rowcolor{mycol1}
OmniSelect (Ours) & 30\% & \textbf{12.47} & 164 ($3.62\times$) & \textbf{2.64 ($1.19\times$)} & \textbf{44.42} \\
\midrule
\multicolumn{6}{c}{\emph{Qwen2.5-Omni-7B}} \\
\midrule
\rowcolor{mycol3}
\multicolumn{6}{l}{\textit{Frame Budgets 128:}} \\
\rowcolor{mycol2!40!white}
Full Tokens & 100\% & 25.08 & 1119 ($1.00\times$) & 3.66 ($1.00\times$) & 46.82 \\
DyCoke (V\&A) & 50\% & 23.00 & 536 ($2.09\times$) & 3.18 ($1.15\times$) & 43.95 \\
OmniZip & 30\% & 22.31 & 319 ($3.51\times$) & 2.94 ($1.24\times$) & 43.85 \\
\rowcolor{mycol1}
OmniSelect (Ours) & 30\% & \textbf{22.28} & \textbf{317 ($3.53\times$)} & \textbf{2.76 ($1.33\times$)} & \textbf{44.17} \\
\bottomrule
\end{tabular}
}

\label{tab:efficiency}
    \vspace{-20pt}
\end{wraptable}

Moreover, OmniSelect maintains strong robustness under aggressive compression, preserving 94.3\%--97.4\% of the original model accuracy on the WorldSense benchmark, while OmniZip and DyCoke exhibit larger performance degradation. These results demonstrate that OmniSelect achieves a favorable trade-off between inference efficiency and reasoning quality.

\subsection{Ablation Study}
\label{sec:4.4}

\textbf{Ablation Study of Threshold $\theta$.} 
The threshold $\theta$ controls the transition between modality-centric pruning and Uniform pruning. To study its effect, we evaluate different fixed $\theta$ values on the WorldSense benchmark when compressing 70\% tokens. As shown in Figure~\ref{fig:threshold}, the extreme cases correspond to pure modality-centric pruning when $\theta \le 0$ and pure Uniform pruning when $\theta \ge 5$. Both Qwen2.5-Omni-3B and 7B exhibit a clear U-shaped performance trend as $\theta$ varies. Specifically, modality-centric pruning at $\theta = 0$ achieves 41.93\% and 42.15\% accuracy for the 3B and 7B models, while Uniform pruning at $\theta = 5$ improves performance to 42.18\% and 42.75\%, respectively. In contrast, intermediate thresholds lead to noticeable degradation, with the 7B model dropping to 41.61\% around $\theta = 2$, which suggests that a fixed intermediate threshold cannot reliably balance modality importance across diverse samples. This indicates that a single fixed threshold is suboptimal and motivates our piecewise design in Section~\ref{sec:4.1}, where $\theta$ is set to $0$ for small modality gaps and $5$ for large gaps to better balance modality importance across different samples.

\textbf{The Significance of Pruning Strategy Classification.} To validate the necessity of nuanced token selection, we evaluate the performance of OmniSelect against three baseline strategies: Audio-Centric, Video-Centric, and Uniform pruning. Experiments are conducted on the DaliyOmni benchmark across two models under two aggressive compression constraints (retaining 30\% and 45\% of audio and video tokens).

\begin{wrapfigure}{r}{0.60\textwidth}
    \centering
    \vspace{-20pt}
    \includegraphics[width=0.60\textwidth,trim=0 0 0 0, clip]{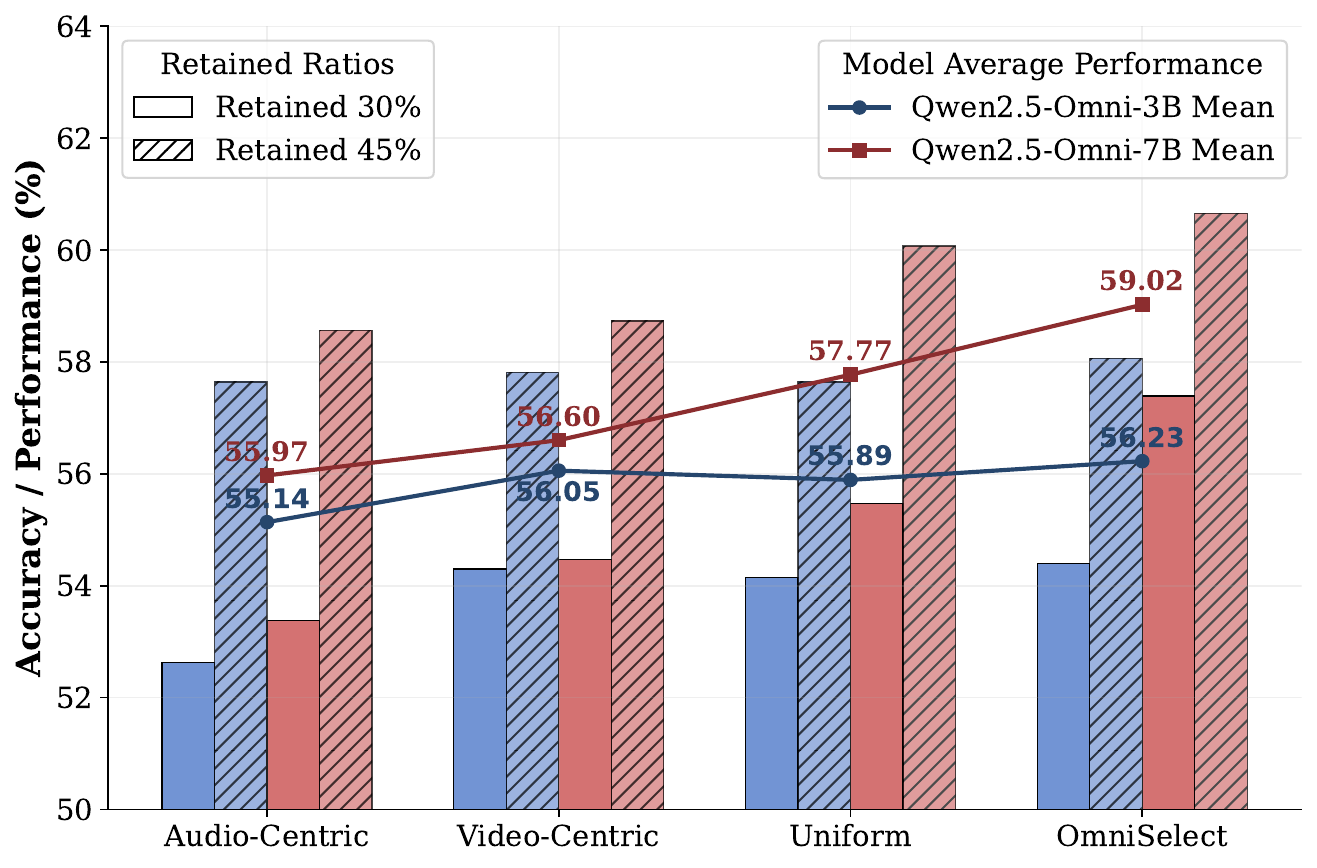}
    \caption{Comparison of different token compression strategies under 30\% and 45\% retained ratios on Qwen-2.5-Omni-3B and Qwen-2.5-Omni-7B. OmniSelect consistently achieves the best performance across all settings.}
    \vspace{-10pt}
    \label{fig:method_compare}
\end{wrapfigure}

As illustrated in Figure~\ref{fig:method_compare}, single-dimensional strategies exhibit clear limitations. For instance, at a 70\% pruning rate, Audio-Centric pruning often discards vital visual context, while Uniform pruning fails to prioritize modality-specific information density, resulting in suboptimal accuracies of 54.14\% and 55.47\% respectively. Even when the retention budget increases to 45\%, these fixed heuristics struggle to match the adaptive allocation of OmniSelect. Notably, OmniSelect achieves a peak performance of 60.65\% on the 7B model, consistently outperforming the best-performing baseline. These results demonstrate the importance of our proposed taxonomy. Relying on any single strategy alone is insufficient to robustly handle diverse scenarios; instead, integrating all three categories is necessary to achieve consistently superior performance.

\textbf{Asymmetric Pruning Analysis.}
We study the effect of modality-specific pruning operators on WorldSense at a 30\% retention ratio. As shown in Table~\ref{tab:asymmetric_pruning}, our asymmetric design, which combines similarity-based video pruning ($V_{\mathrm{sim}}$) with attention-based audio pruning ($A_{\mathrm{attn}}$), achieves the highest overall accuracy of 44.42\%. It outperforms both the Similarity Only variant ($V_{\mathrm{sim}}+A_{\mathrm{sim}}$, 43.25\%) and the Attention Only variant ($V_{\mathrm{attn}}+A_{\mathrm{attn}}$, 40.70\%).
\begin{wraptable}{r}{0.60\columnwidth}
\captionsetup{
    skip=4pt,
    width=\linewidth,
    justification=raggedright,
    singlelinecheck=false
}
\renewcommand{\arraystretch}{0.94}
\setlength{\tabcolsep}{4pt}

\caption{\textbf{Asymmetric Pruning Analysis on WorldSense.} Accuracy (\%) at a 30\% retention ratio. The \textbf{bold} number in each row indicates the best result.}
\label{tab:asymmetric_pruning}

\resizebox{\linewidth}{!}{
\begin{tabular}{lccc}
\toprule
Category
& \makecell{Ours\\$V_{\mathrm{sim}}+A_{\mathrm{attn}}$}
& \makecell{Similarity Only\\$V_{\mathrm{sim}}+A_{\mathrm{sim}}$}
& \makecell{Attention Only\\$V_{\mathrm{attn}}+A_{\mathrm{attn}}$} \\
\midrule
Overall       & \textbf{44.42} & 43.25 & 40.70 \\
Music         & \textbf{46.80} & 43.84 & 44.33 \\
Games         & \textbf{41.20} & 39.48 & 34.33 \\
Tech./Science & \textbf{49.80} & 48.37 & 47.14 \\
Daily Life    & \textbf{43.92} & 41.95 & 40.58 \\
Film/TV       & \textbf{43.01} & 41.69 & 39.58 \\
Sports        & \textbf{39.53} & 39.07 & 35.81 \\
Culture/Pol.  & 47.57 & \textbf{48.22} & 41.75 \\
Performance   & 41.20 & \textbf{42.70} & 37.45 \\
\bottomrule
\end{tabular}
}
\end{wraptable}
Furthermore, keeping audio pruning fixed and replacing $V_{\mathrm{sim}}$ with $V_{\mathrm{attn}}$ reduces the accuracy of the Games category by 6.87 points, from 41.20\% to 34.33\%. This result suggests that similarity-based video pruning is more effective for removing temporally redundant visual tokens. Conversely, keeping video pruning fixed and replacing $A_{\mathrm{attn}}$ with $A_{\mathrm{sim}}$ reduces the accuracy of Music category by 2.96 points, from 46.80\% to 43.84\%. This indicates that attention-based audio pruning better preserves salient acoustic information in audio-sensitive tasks.

We further performed a latency spot check on 20 randomly selected samples. The attention-based audio pruning variant was faster on 13 samples, corresponding to 65\% of the cases, with an average latency reduction of 11.2\,ms and a maximum reduction of 134.4\,ms. This result is consistent with the lower computation overhead of reusing attention-derived saliency, while similarity-based audio pruning requires additional pairwise feature similarity computation. Overall, the results support the use of modality-specific pruning operators and provide empirical evidence for our asymmetric design.

\section{Conclusion}
 This paper presents OmniSelect, a training-free, modality-adaptive token compression framework for Omni-LLMs. 
Specifically, the framework dynamically selects appropriate compression strategies for multimodal inputs and performs fine-grained token pruning within each temporal group, adaptively allocating pruning ratios to preserve informative tokens across modalities. 
Extensive experimental results on audio-visual understanding tasks with two OmniLLMs (3B and 7B parameters) demonstrate that OmniSelect achieves competitive performance among existing training-free token compression approaches.
OmniSelect achieves an inference speedup of 1.19$\times$ to 1.33$\times$ while reducing GPU memory consumption by 2.58GB to 2.77GB. Despite these substantial resource savings, OmniSelect retains 94\% to 99\% of the original full-token accuracy.

\section{Acknowledgments}
We want to thank all the anonymous reviewers for their valuable comments. This work was supported by the National Natural Science Foundation of China (NSFC No.  62622608 and 62576232).

\bibliography{custom}
\bibliographystyle{plain}


\newpage

\appendix
\section{Preliminary}
\label{sec: preliminary}
An OmniLLM \citep{jiang2025specificmllmsomnimllmssurveymllms} typically consists of a vision encoder \(\mathcal{E}_{v}\), an audio encoder \(\mathcal{E}_{a}\), and an LLM backbone. In the input pipeline, video, audio, and textual prompts are treated as raw data. Specifically, video sequences are converted into frames \(V\) via uniform temporal sampling, while audio signals \(A\) are processed using a fixed sampling rate. The inputs of the two modalities are mapped into token sequences as follows:
\begin{equation}
\mathcal{T}_{v} = \mathcal{P}_{v}(\mathcal{E}_{v}(V)), \quad \mathcal{T}_{a} = \mathcal{P}_{a}(\mathcal{E}_{a}(A)),
\end{equation}
where \(\mathcal{T}_{v} \in \mathbb{R}^{N_{v} \times D}\) and \(\mathcal{T}_{a} \in \mathbb{R}^{N_{a} \times D}\) represent the token embedding sequences of visual and audio modalities. Here, \(N_v\) and \(N_a\) denote the number of visual and audio tokens, and \(D\) is the hidden dimension of the LLM backbone. The projectors \(\mathcal{P}_v\) and \(\mathcal{P}_a\) align the encoded features from their respective latent spaces into the joint embedding space of the LLM backbone.

Moreover, to achieve temporal alignment between visual and auditory modalities, OmniLLMs often employ a token-level concatenation strategy within synchronized time windows. Specifically, the original token sequences are partitioned into \(K\) temporal segments, where the visual and audio tokens within the same time interval are concatenated to form joint multimodal embeddings. This process can be formally defined as:
\begin{equation}
\mathcal{T}_{m}^{(i)} = [ \mathcal{T}_{v}^{(i)} \parallel \mathcal{T}_{a}^{(i)} ], \quad i = 1, 2, \dots, K
\end{equation}
where \(\mathcal{T}_{v}^{(i)}\) and \(\mathcal{T}_{a}^{(i)}\) denote the \(i\)-th segments of visual and audio tokens respectively, and \(\parallel\) represents the concatenation operation. Consequently, the resulting sequence \(\hat{\mathcal{T}} = \{ \mathcal{T}_{m}^{(1)}, \mathcal{T}_{m}^{(2)}, \dots, \mathcal{T}_{m}^{(K)} \}\) encapsulates synchronized bimodal information within each unified token. This enables the LLM backbone to model cross-modal dependencies in a temporally coherent latent space. These chunked multimodal tokens are subsequently concatenated with the textual embeddings \(\mathcal{T}_t\) to form the unified input sequence.

\begin{algorithm}[t]
\caption{OmniSelect: Dynamic Modality-Aware Token Compression}
\label{alg:omniselect}

\textbf{Input:} Video tokens $T_v$, audio tokens $T_a$, query $q$, pruning ratio $\rho_{v/a}$, number of Frames $F$ \\
\textbf{Output:} Pruned multimodal tokens $\hat{T}$

\begin{algorithmic}[1]

\State Divide $(T_v,T_a)$ into $G$ aligned temporal groups

\State Compute AudioCLIP embeddings:
$v_i \gets \text{AudioCLIP}(T_v^i)$,
$a_i \gets \text{AudioCLIP}(T_a^i)$,
$t \gets \text{AudioCLIP}(q)$

\For{$i=1$ to $F$}
    \State $s_i^{(v)} \gets \cos(v_i,t)$,
    $s_i^{(a)} \gets \cos(a_i,t)$
\EndFor

\State $\bar{s}^{(v)} \gets \frac{1}{F}\sum_i s_i^{(v)}$,
$\bar{s}^{(a)} \gets \frac{1}{F}\sum_i s_i^{(a)}$

\If{$|\bar{s}^{(v)}-\bar{s}^{(a)}| \le \theta$}
    \State Strategy $\gets$ Uniform
\ElsIf{$\bar{s}^{(v)} > \bar{s}^{(a)}$}
    \State Strategy $\gets$ Video-Centric
\Else
    \State Strategy $\gets$ Audio-Centric
\EndIf

\For{$i=1$ to $G$}

    \State $\rho_i \gets
    \sigma\!\left(
    -\frac{\hat{s}_i-\mu_s}{\tau\sigma_s}
    \right)$

    \State $K_{i,a/v} \gets \lfloor (1-\rho_i)\cdot n_i \rfloor \text{ or } \lfloor(1 - \rho_{v/a}) \cdot n_i\rfloor$

    \State $A \gets \text{Softmax}\!\left(\frac{QK^\top}{\sqrt{d}}\right)$ \Comment{global attention computed once}

    \State $A_i \gets A[\mathcal{G}_i]$ \Comment{slice attention for group $i$}

    \State $I_a^i \gets \text{TopK}(A_i,K_{i,a})$

    \State $S_i \gets \bar{V}_i\bar{V}_i^\top$

    \State $I_v^i \gets \text{BottomK}(S_i,K_{i,v})$

    \State $\hat{T}_i \gets
    T_v^i[I_v^i]\cup T_a^i[I_a^i]$

\EndFor

\State $\hat{T} \gets \{\hat{T}_1,\hat{T}_2,\dots,\hat{T}_G\}$

\State \Return $\hat{T}$

\end{algorithmic}
\end{algorithm}

\begin{figure*}[!hbtp]
    \centering
    \begin{subfigure}[b]{0.50\textwidth}
        \centering
        \includegraphics[width=\textwidth,trim=250 150 250 140,clip]{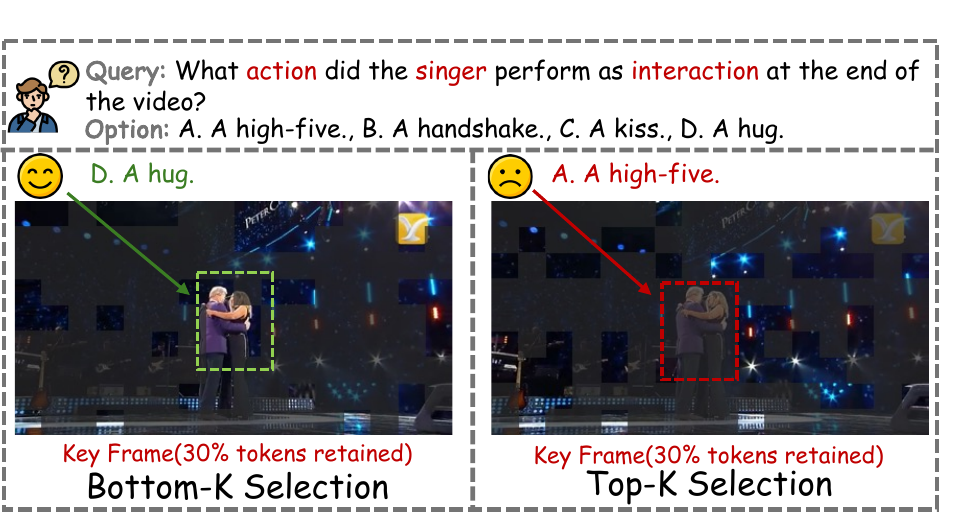}
        \caption{Bottom-K Selection vs. Top-K Selection.}
        \label{fig:vs}
    \end{subfigure}
    \hfill
    \begin{subfigure}[b]{0.49\textwidth}
        \centering
        \includegraphics[width=\textwidth,trim=100 70 120 55,clip]{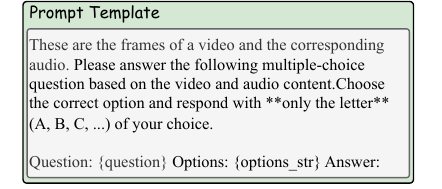}
        \label{fig:prompt}
        \caption{Prompt Template.}
    \end{subfigure}

    \caption{\textbf{Left:} The comparison of Bottom-K Selection strategy and Top-K Selection strategy. \textbf{Right:} Prompt Template for multiple-choice QA evaluations.}
    \label{fig:prompt}
\end{figure*}
\section{Why Bottom-K Strategy Work for Video Token Pruning?}

We adopt video embeddings instead of attention weights for two main reasons. First, computing full attention maps over a large number of vision tokens incurs prohibitive computational overhead, whereas embedding-based similarity is significantly more efficient and scalable. Second, feature-level similarity offers a more direct and stable measure of token redundancy, effectively avoiding the attention sink bias and noise commonly observed in pre-trained attention heads.

Furthermore, pruning is performed within each temporal group to meet the pre-allocated token budget defined in Section~\ref{sec: modalityaware}. Thanks to the dynamic proportion allocation mechanism, which already mitigates a substantial portion of redundancy across the temporal dimension by assigning higher token budgets to more informative segments, the remaining redundancy becomes predominantly localized within each temporal group. As illustrated in Figure~\ref{fig:vs}, this makes the Bottom-K strategy particularly effective: informative visual cues are typically sparsely distributed over time, while most tokens inside a local temporal window remain highly redundant.

By selecting tokens with the lowest similarity scores, Bottom-K preferentially removes redundant or highly correlated content while preserving visually distinctive tokens that are more likely to align with textual semantics. As shown in the figure, when answering “What action did the singer perform as interaction at the end of the video?”, the Bottom-K selection accurately captures the critical hugging scene (correct answer: D. A hug), whereas Top-K focuses on a less relevant region and leads to the wrong prediction (A. A high-five). This approach achieves more balanced temporal coverage and effectively mitigates information collapse within dense regions. Overall, the Bottom-K strategy provides a simple yet powerful mechanism to retain representative and semantically diverse visual tokens, which is crucial for maintaining strong reasoning performance under aggressive compression.

\begin{table*}[t]
\centering
\captionsetup{skip=4pt}
\renewcommand{\arraystretch}{0.94} %
\setlength{\tabcolsep}{4pt} %
\caption{{
\textbf{WorldSense Results under Any-Correct Evaluation with Strategy Diversity.} \textbf{*} denotes performance exceeding Full Tokens. The best results are \textbf{bolded} for each column.
}
}
\resizebox{1\linewidth}{!}{
\begin{tabular}{ccccccccccc}
\toprule
Method & Retained Ratio & \begin{tabular}[c]{@{}c@{}}Tech \&\\ Science\end{tabular} & Games & \begin{tabular}[c]{@{}c@{}}Daily\\ Life\end{tabular} & \begin{tabular}[c]{@{}c@{}}Film \&\\ TV\end{tabular} & Music & Sports & \begin{tabular}[c]{@{}c@{}}Culture \&\\ Politics\end{tabular} & Performance & Avg. ($\uparrow$) \\ \midrule
\multicolumn{11}{c}{\emph{Qwen2.5-Omni-3B}} 
\\ 
\midrule
\rowcolor{mycol3}
\multicolumn{11}{l}{\textit{Frame Budgets 128:}} \\
\rowcolor{mycol2!40!white}
Full Tokens & 100\% & \textbf{52.65} & 39.06 & 43.47 & 44.06 & 46.55 & 42.79 & 52.43 & 41.20 & 45.62 \\
\rowcolor{mycol1}
OmniSelect (Ours) & 45\% & 51.63 & \textbf{43.78}$^{*}$ & {45.14}$^{*}$ & \textbf{44.33}$^{*}$ &  48.03$^{*}$ & \textbf{43.72}$^{*}$ &  \textbf{53.07}$^{*}$ & 41.57$^{*}$ & \textbf{46.60}$^{*}$ \\ 
\rowcolor{mycol1}
OmniSelect (Ours) & 30\% & 51.43 & 42.92$^{*}$ & \textbf{45.59}$^{*}$ &  43.27 &  \textbf{48.52}$^{*}$ & 41.63 & {50.81} & \textbf{42.70}$^{*}$ &  46.12$^{*}$ \\
\midrule
\multicolumn{11}{c}{\emph{Qwen2.5-Omni-7B}} 
\\ 
\midrule
\rowcolor{mycol3}
\multicolumn{11}{l}{\textit{Frame Budgets 128:}} \\
\rowcolor{mycol2!40!white}
Full Tokens & 100\% & \textbf{49.39} & 42.49 & \textbf{46.96} & 44.59 & 48.03 & 42.56 & \textbf{53.72} & \textbf{45.69} & 46.82 \\
\rowcolor{mycol1}
OmniSelect (Ours) & 45\% & 48.78 & \textbf{43.35}$^{*}$ & 46.81 & \textbf{45.38}$^{*}$ & \textbf{49.26}$^{*}$ & 44.19$^{*}$ & 51.78 &  \textbf{45.69} & \textbf{47.04}$^{*}$ \\
\rowcolor{mycol1}
OmniSelect (Ours) & 30\%  & 47.96 & 42.06 & 46.50 & {44.85}$^{*}$ & 46.80 &  \textbf{46.05}$^{*}$&  49.84 &  44.57 & 46.34 \\
\bottomrule
\end{tabular}
}
\label{tab:wb}
\vspace{-3mm}
\end{table*}

\begin{table*}[t]
\centering
\captionsetup{skip=4pt}
\renewcommand{\arraystretch}{0.94} %
\setlength{\tabcolsep}{4pt} %
\caption{{
\textbf{DailyOmni Results under Any-Correct Evaluation with Strategy Diversity.} \textbf{*} denotes performance exceeding Full Tokens. The best results are \textbf{bolded} for each column.
}
}
\resizebox{1\linewidth}{!}{
\begin{tabular}{ccccccccc}
\toprule
Method & Retained Ratio & Con. & Event & AV Event & Com. & Inf. & Rea. & Avg. ($\uparrow$) \\ \midrule
\multicolumn{9}{c}{\emph{Qwen2.5-Omni-3B}} 
\\ 
\midrule
\rowcolor{mycol3}
\multicolumn{9}{l}{\textit{Frame Budgets 128:}} \\
\rowcolor{mycol2!40!white}
Full Tokens & 100\% & 55.44 & \textbf{56.21} &  53.78 & \textbf{70.99} & 79.22 & \textbf{74.29} & \textbf{62.82} \\
\rowcolor{mycol1}
OmniSelect (Ours) & 45\% & 54.92 & 54.25 & {57.14}$^{*}$ & 67.94 &  \textbf{79.87}$^{*}$ & 72.00 &  62.32\\ 
\rowcolor{mycol1}
OmniSelect (Ours) & 30\% & \textbf{55.96}$^{*}$ & 52.94$^{*}$ & \textbf{57.56}$^{*}$ & 67.18 &  75.32 & 68.57 & {61.07}\\
\midrule
\multicolumn{9}{c}{\emph{Qwen2.5-Omni-7B}} 
\\ 
\midrule
\rowcolor{mycol3}
\multicolumn{9}{l}{\textit{Frame Budgets 128:}} \\
\rowcolor{mycol2!40!white}
Full Tokens & 100\% & \textbf{56.99} & \textbf{60.13} & \textbf{50.84} &  71.76 & 78.57 & \textbf{78.86} & \textbf{64.16} \\
\rowcolor{mycol1}
OmniSelect (Ours) & 45\% & 54.92 & 54.90 & \textbf{50.84} & \textbf{74.05}$^{*}$ & 77.27 & 76.57 & 62.24\\
\rowcolor{mycol1}
OmniSelect (Ours) & 30\%  & 52.33 & 53.59 & 49.16 & {69.47} & \textbf{79.22}$^{*}$ &  74.29 &  60.57 \\
\bottomrule
\end{tabular}
}
\label{tab:dailyomni_best}
\vspace{-3mm}
\end{table*}

\section{Dynamic Modality-Aware Token Compression Algorithm}

To clearly present our overall pipeline, we summarize the full dynamic modality-aware token compression process in Algorithm~\ref{alg:omniselect}. The algorithm first aligns video and audio tokens into $G$ temporal groups to establish fine-grained cross-modal correspondence. It then computes AudioCLIP-based embeddings for each modality and the query, and estimates group-wise semantic similarities to determine the relative relevance of video and audio signals. Based on the aggregated similarity, a dynamic strategy is selected (video-centric, audio-centric, or uniform) to adaptively balance modality importance. Finally, under the chosen strategy, we perform group-wise token pruning with adaptive retention ratios and attention-guided selection, and merge the retained audio and video tokens to form the final compressed multimodal representation $\hat{T}$.

\begin{figure}[t]
    \centering
    \begin{subfigure}[b]{0.49\textwidth}
        \centering
        \includegraphics[width=1.0\textwidth,trim=5 6 5 5, clip]{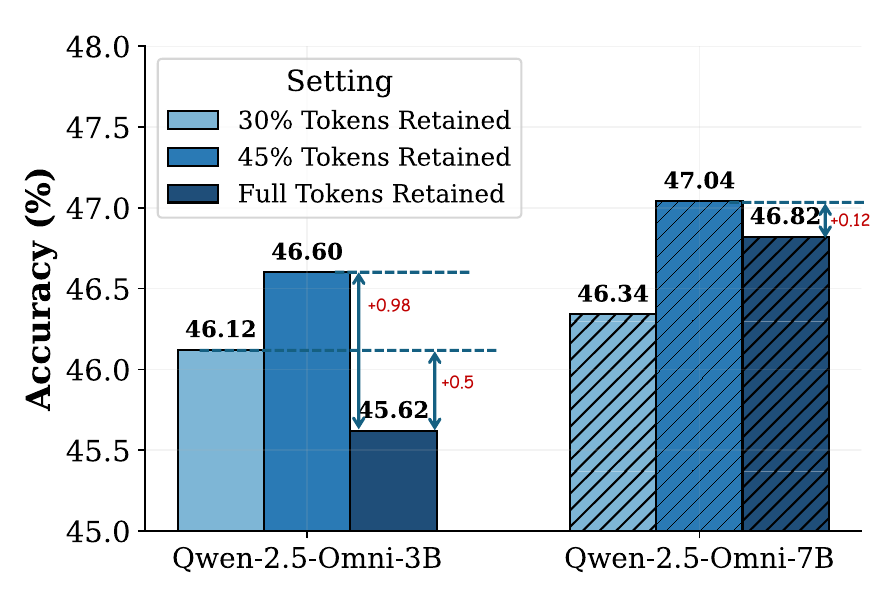}
        \label{fig:best_results_worldsense}
    \end{subfigure}
    \hfill
    \begin{subfigure}[b]{0.49\textwidth}
        \centering
        \includegraphics[width=\textwidth,trim=5 6 5 5, clip]{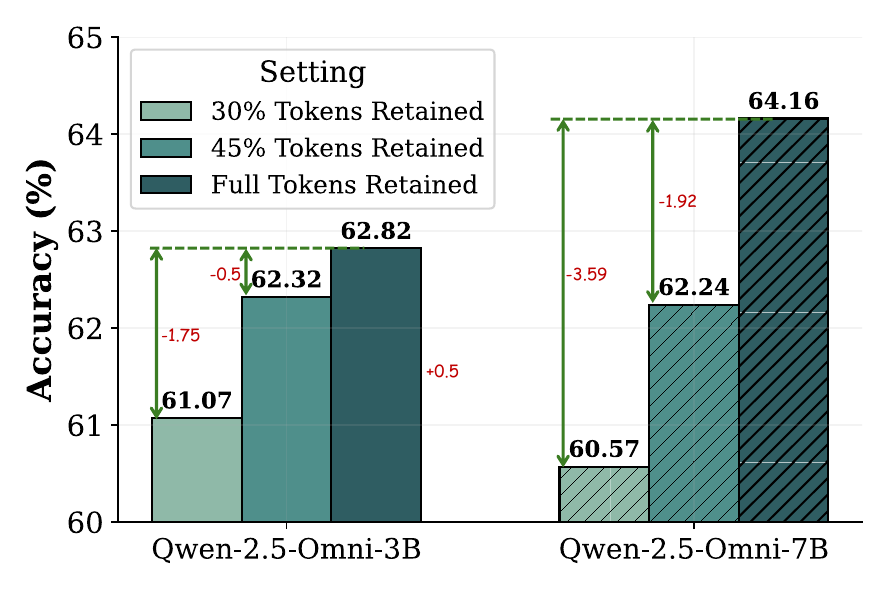}  
        \label{fig:best_results_dailyomni}
    \end{subfigure}
    
    \caption{Qwen-2.5-Omni-3B and Qwen-2.5-Omni-7B performance on WorldSense (\textbf{Left}) and DailyOmni (\textbf{Right}) benchmarks at 30\% and 45\% pruning ratios.}
    \label{fig:bestresults}
\end{figure}

\section{More Experimental Details}

\subsection{More Experimental Results}
\label{more exp results}

\begin{figure}[t]
    \centering
    \begin{subfigure}[b]{0.49\textwidth}
        \centering
        \includegraphics[width=1.0\textwidth,trim=5 6 5 5, clip]{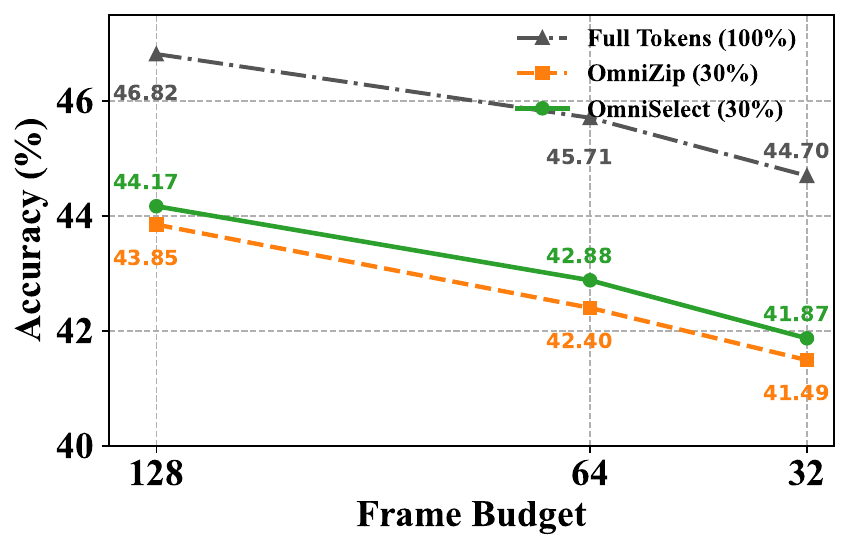}
        \label{fig:budget_30_7b}
    \end{subfigure}
    \hfill
    \begin{subfigure}[b]{0.49\textwidth}
        \centering
        \includegraphics[width=\textwidth,trim=5 6 5 5, clip]{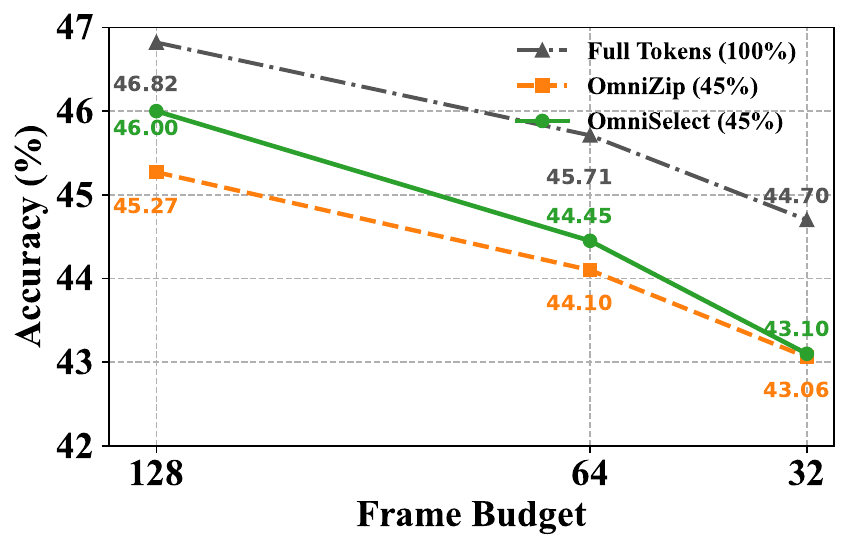}  
        \label{fig:budget_45_7b}
    \end{subfigure}
    
    \caption{Qwen-2.5-Omni-7B performance under varying frame budgets at 30\% and 45\% pruning.}
    \label{fig:budgets_7b}
\end{figure}

 We also evaluate the Qwen-2.5-Omni-7B model on the WorldSense benchmark under frame budgets of 32 and 64. As shown in Table~\ref{tab:WorldSense_7b}, under the same compression settings, OmniSelect consistently outperforms OmniZip across different frame budgets. Furthermore, as illustrated in Figure~\ref{fig:budgets_7b}, although the performance gains brought by token pruning on Qwen-2.5-Omni-7B are less pronounced than those observed on the 3B model, OmniSelect still consistently achieves better performance than OmniZip under the same compression ratio across all three frame budget settings. These results further demonstrate the robustness and effectiveness of our modality-aware dynamic pruning strategy on larger OmniLLMs.

\subsection{More Implementation Details}
\label{more imp details}

\textbf{Prompt for QA Evaluation.} 
For all QA-based benchmarks, we use a unified prompt template to ensure consistent evaluation across different datasets. Specifically, each input is formatted as shown in Figure~\ref{fig:prompt}. This simple and instruction-consistent template is used for all QA evaluations without task-specific modifications.

\textbf{GPUs.} 
All experiments are conducted on NVIDIA A100 GPUs with 40GB memory. For most benchmarks, a single GPU is sufficient to run inference. For VideoMME, due to its longer video duration and higher computational load, we use two GPUs to ensure efficient processing, while all other tasks are evaluated on a single GPU.

\textbf{Input Configuration and Preprocessing.}
For inference, video inputs are uniformly sampled at a rate of 2 frames per second (FPS), with the total frame count restricted to a maximum of 32 / 64 / 128 / 512. Following the setting of baselines for fair comparison, the spatial resolution for each individual frame is configured at a maximum of $128 \times 28 \times 28$ pixels.

\begin{table*}[t]
\centering
\captionsetup{skip=4pt}
\renewcommand{\arraystretch}{0.94} %
\setlength{\tabcolsep}{4pt} %
\caption{{
\textbf{WorldSense results when the frame budgets are set to 32 and 64.} \textbf{*} denotes performance exceeding Full Tokens. The best and second-best results are \textbf{bolded} and \textbf{underlined} for each column, respectively.
}
}
\resizebox{1\linewidth}{!}{
\begin{tabular}{ccccccccccc}
\toprule
Method & Retained Ratio & \begin{tabular}[c]{@{}c@{}}Tech \&\\ Science\end{tabular} & Games & \begin{tabular}[c]{@{}c@{}}Daily\\ Life\end{tabular} & \begin{tabular}[c]{@{}c@{}}Film \&\\ TV\end{tabular} & Music & Sports & \begin{tabular}[c]{@{}c@{}}Culture \&\\ Politics\end{tabular} & Performance & Avg. ($\uparrow$) \\ \midrule

\multicolumn{11}{c}{\emph{Qwen2.5-Omni-7B}} 
\\ 
\midrule
\rowcolor{mycol3}
\multicolumn{11}{l}{\textit{Frame Budgets 32:}} \\
\rowcolor{mycol2!40!white}
Full Tokens & 100\% & 47.55 & 41.63 & 44.68 & 42.74 & 46.31 & 41.40 & 49.51 & 42.32 & 44.70 \\
OmniZip & 45\%  &  \textbf{45.92} & \underline{39.91} & \textbf{43.77} & 37.99 & \textbf{47.04} & \textbf{41.40} & \underline{45.31} &  \underline{40.07} & \underline{43.06}  \\
OmniZip & 30\%  &  \underline{45.10} & \textbf{40.34} & 40.43 & \underline{38.52} & \underline{46.80} & 38.60 & 42.07 &  38.58 & 41.49 \\
\rowcolor{mycol1}
OmniSelect (Ours) & 45\% & 44.90 & \underline{39.91} & \textbf{43.62} &  \textbf{39.05} & 46.31 & 39.77 & \textbf{47.90} &  \textbf{41.95} & \textbf{43.10} \\
\rowcolor{mycol1}
OmniSelect (Ours) & 30\%  & \underline{45.10} & 36.91 & 41.34 & \underline{38.52} & 45.32 &  \underline{41.16} &  44.66 &  38.95 & 41.87 \\
\midrule
\rowcolor{mycol3}
\multicolumn{11}{l}{\textit{Frame Budgets 64:}} \\
\rowcolor{mycol2!40!white}
Full Tokens & 100\% & 48.78 & 41.20 & 45.39 & 44.85 & 47.04 & 41.16 & 51.46 & 44.19 & 45.71 \\
OmniZip & 45\% & \textbf{47.14} & \underline{39.48} & \textbf{44.53} & 40.63 & \underline{47.54}$^{*}$ & 40.93 & \underline{46.93} & \underline{43.07} &  \underline{44.10} \\
OmniZip & 30\%  &  45.10 & 37.34 & 41.49 & \underline{40.90} & 46.80 & 40.47 & 44.34 &  40.45 & 42.40 \\
\rowcolor{mycol1}
OmniSelect (Ours) & 45\% & \underline{45.51} & \textbf{41.63}$^{*}$ & \underline{43.77} &  \textbf{42.22} & \textbf{48.28}$^{*}$ & \textbf{41.86}$^{*}$ & \textbf{47.90} &  \textbf{44.19} & \textbf{44.45} \\
\rowcolor{mycol1}
OmniSelect (Ours) & 30\%  & \underline{45.51} & 38.20 & 43.31 & {39.84} & 46.80 &  \underline{41.16} &  44.34 &  40.45 & 42.88 \\
\bottomrule
\end{tabular}
}
\label{tab:WorldSense_7b}
\vspace{-3mm}
\end{table*}

\subsection{Any-Correct Evaluation under Strategy Diversity}
\label{best results}

To further analyze the robustness of our modality-aware strategy selection, we introduce a relaxed evaluation setting that considers all three candidate pruning strategies, including Uniform, Video-Centric, and Audio-Centric Pruning. In this setting, a sample is regarded as correctly solved if any one of the three strategies produces a correct prediction, regardless of the selected strategy.

We observe that the performance gap under the threshold-based strategy partition ($\theta$) mainly stems from the limited discriminative capability of AudioCLIP in certain challenging cases, which may lead to suboptimal modality preference estimation. Since inference can only be performed once in practical settings, the strategy must be determined in a single forward pass, making the threshold-based decision inherently sensitive to such estimation errors. As shown in Table~\ref{tab:wb} and Table~\ref{tab:dailyomni_best}, this relaxed setting achieves performance very close to the Full Tokens setting, with only marginal degradation and, in some cases, even higher performance. For instance, as shown in Figure~\ref{fig:bestresults}, when the retained ratio is set to 45\%, OmniSelect improves the accuracy of Qwen-2.5-Omni-3B and Qwen-2.5-Omni-7B by 0.98\% and 0.12\%, respectively. Meanwhile, on the DailyOmni benchmark, it consistently preserves 94.4\%--99.2\% of the Full Tokens' performance, substantially outperforming threshold-based control methods.  This indicates that the optimal strategy often exists within the proposed strategy space, and the main limitation lies in strategy selection rather than representation capacity.

Overall, these results further demonstrate that our dynamic modality-aware strategy is both reasonable and effective in balancing compression and performance, while also highlighting a potential direction for future improvement in more robust strategy prediction under weak modality discriminability.

\begin{figure*}[t]
    \centering
    \includegraphics[width=0.8\textwidth]{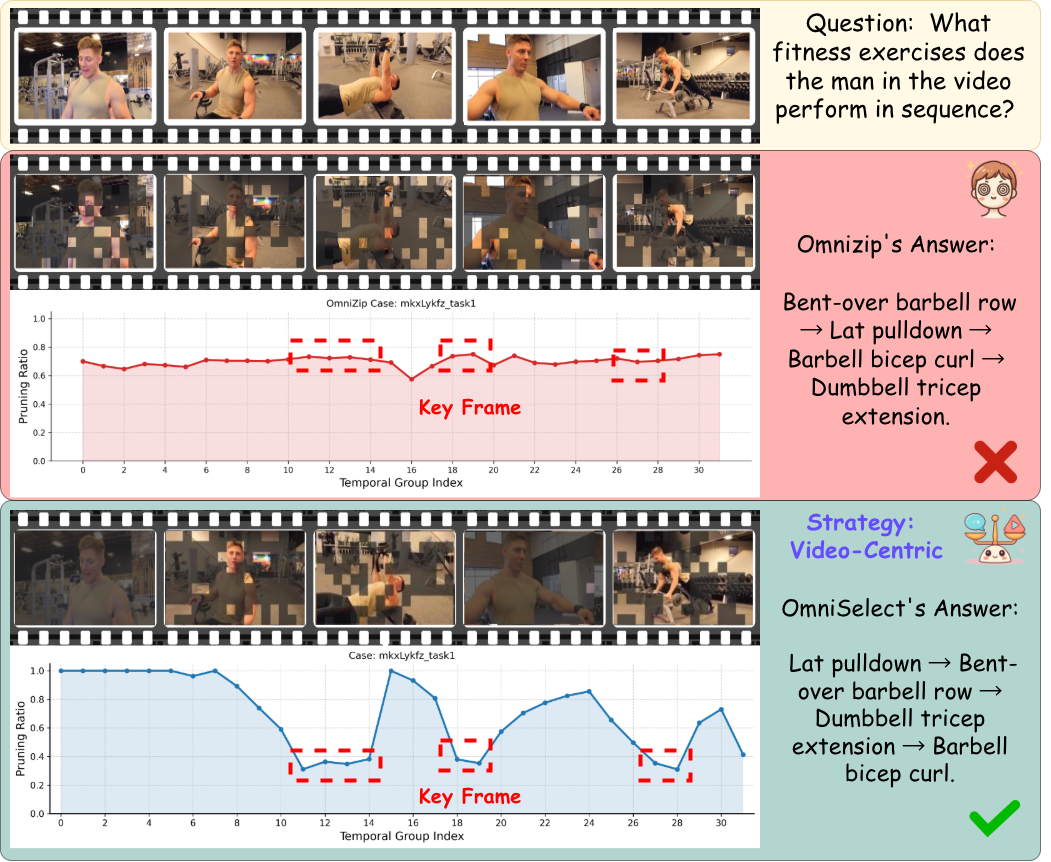}
    \caption{A Video-Centric pruning case, including pruning results, answers, and per-temporal-group pruning ratios for our method and the baseline.}
    \label{fig:case1}
\end{figure*}

\begin{figure*}[t]
    \centering
    \includegraphics[width=0.8\textwidth]{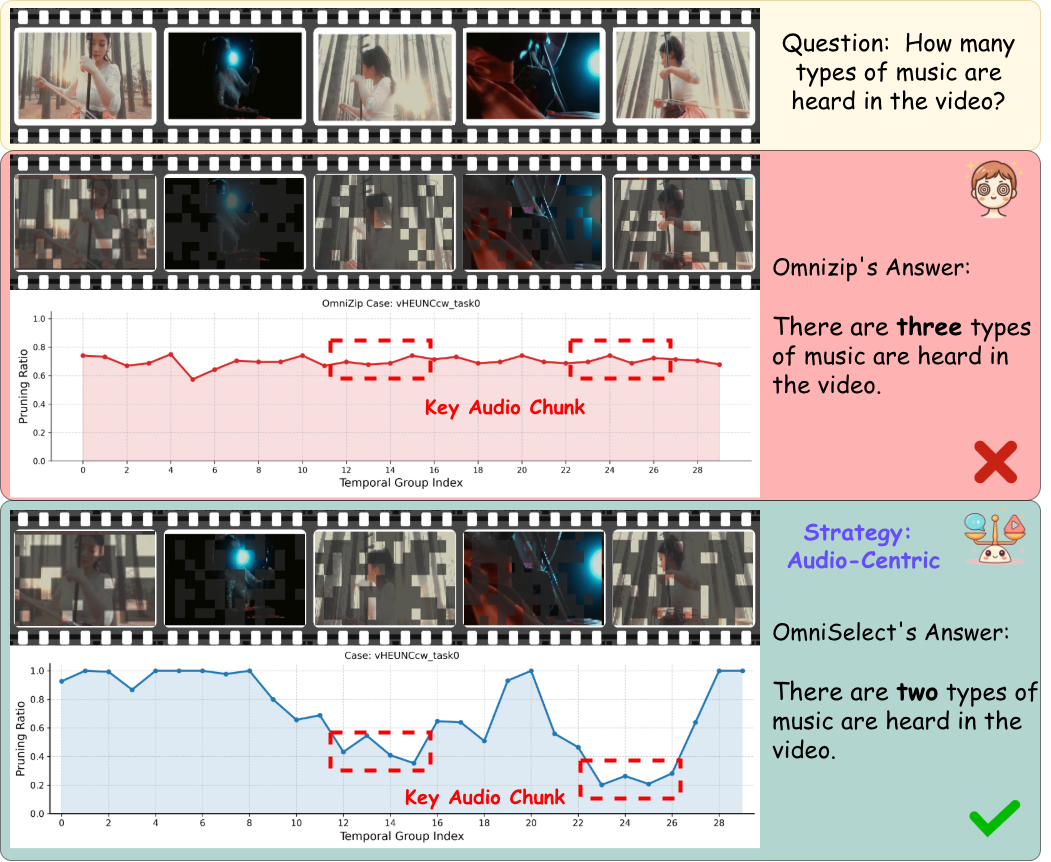}
    \caption{A Audio-Centric pruning case, including pruning results, answers, and per-temporal-group pruning ratios for our method and the baseline.}
    \label{fig:case2}
\end{figure*}

\begin{figure*}[!hbtp]
    \centering
    \includegraphics[width=1.0\textwidth,trim=60 90 60 80, clip]{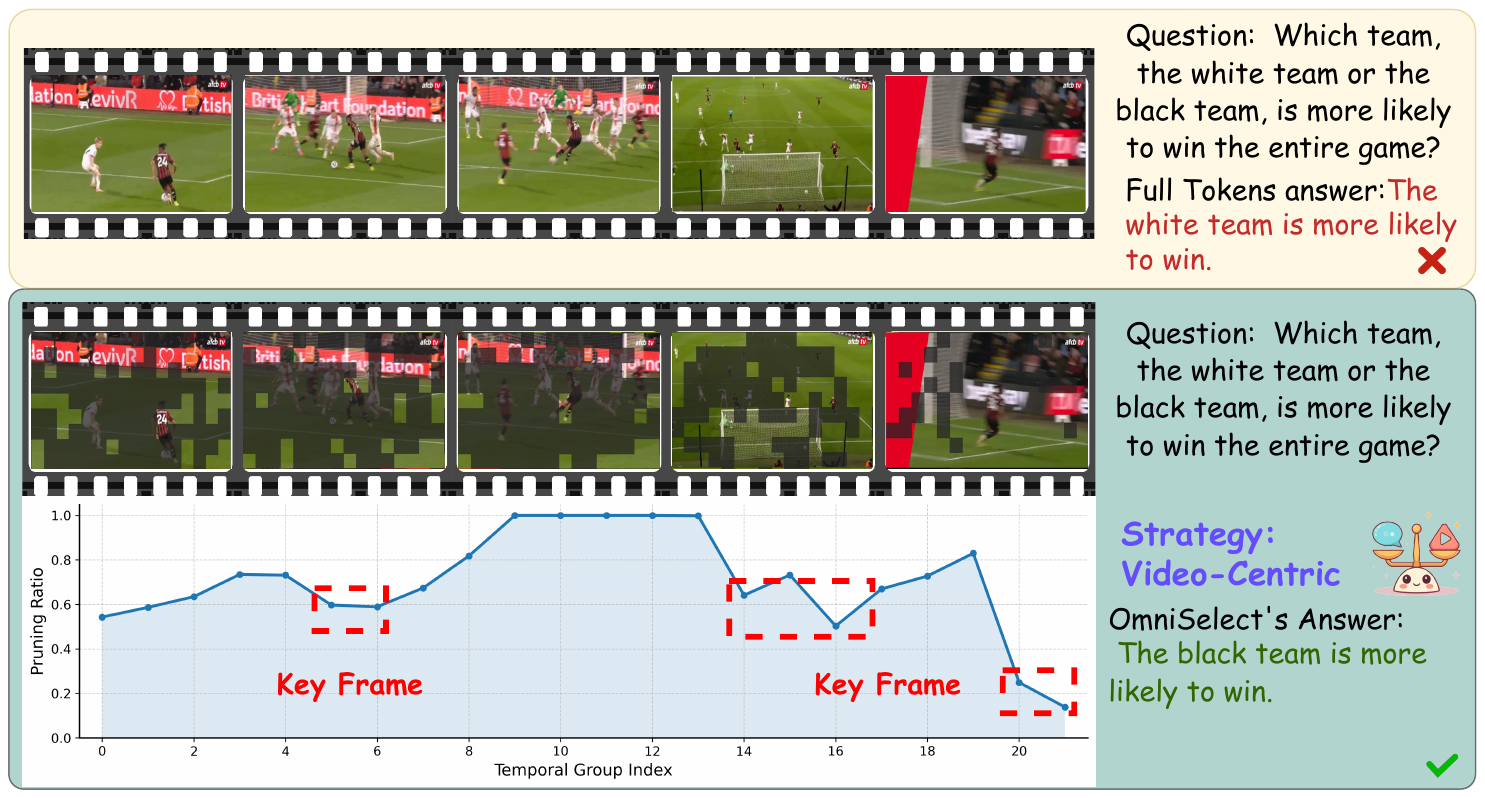}
    \caption{A Video-Centric pruning case that OmniSelect's answer corrects but Full Tokens does not, including pruning results, answers, and per-temporal-group pruning ratios for our method.}
    \label{fig:case3}
\end{figure*}

\section{Relevance Estimator Analysis}
\label{sec:relevance_estimator}

We evaluate the effect of the relevance estimator by replacing AudioCLIP~\citep{guzhov2022audioclip} with ImageBind~\citep{girdhar2023imagebind} under the same experimental setting: WorldSense, Qwen2.5-Omni-7B, 128 input frames, and a 30\% token pruning ratio. 

\begin{table}[!hbtp]
    \centering
    \begin{threeparttable}
        \captionsetup{
            skip=4pt,
            justification=raggedright,
            singlelinecheck=false
        }
        \caption{\textbf{Comparison of relevance estimators.}
        Results are measured on WorldSense using Qwen2.5-Omni-7B,
        128 input frames, and a 30\% token pruning ratio. Best results are \textbf{bolded} in each column.}
        \label{tab:estimator_comparison}

        \small
        \setlength{\tabcolsep}{6pt}
        \renewcommand{\arraystretch}{0.96}
        \begin{tabular}{lcccc}
            \toprule
            Estimator
            & Parameters
            & Accuracy (\%) $\uparrow$
            & Latency (s) $\downarrow$
            & GPU Mem. (GB) $\downarrow$ \\
            \midrule
            AudioCLIP
            & $\sim$200M
            & \textbf{44.17}
            & \textbf{0.191}
            & \textbf{22.28} \\
            ImageBind
            & $\sim$1B
            & 43.88
            & 2.701
            & 26.45 \\
            \bottomrule
        \end{tabular}
    \end{threeparttable}
\end{table}

As shown in Table~\ref{tab:estimator_comparison}, AudioCLIP achieves 44.17\% accuracy, slightly higher than ImageBind at 43.88\%. AudioCLIP also has a substantially smaller model size ($\sim$200M versus $\sim$1B parameters), lower measured latency (0.191\,s versus 2.701\,s), and lower GPU memory consumption (22.28\,GB versus 26.45\,GB). These results indicate that AudioCLIP offers a more favorable accuracy-efficiency trade-off for lightweight, training-free dynamic token selection. We do not include WAVE~\citep{tang2026wave} in this comparison because its 7B parameters substantially exceed the scale of the lightweight relevance estimators considered here. Our goal is to use a training-free estimator with low computational overhead for omni-modal token selection; WAVE, as a large-scale multimodal foundation model, falls outside this comparison scope.

\section{Case Study}

As shown in Figure~\ref{fig:case1}, OmniSelect (our method) adopts a video-centric pruning strategy, aggressively removing tokens from frames that contain little relevant equipment or action information, while preserving more tokens for key fitness-related frames. In contrast, OmniZip (baseline) applies a nearly uniform pruning ratio across all frames regardless of their semantic importance. Consequently, OmniSelect successfully predicts the correct answer, whereas OmniZip fails.

In Figure~\ref{fig:case2}, OmniSelect employs an audio-centric pruning strategy, uniformly pruning visual tokens while dynamically adjusting the pruning ratio for audio tokens, retaining more tokens in acoustically informative segments. By contrast, OmniZip uniformly prunes both visual and audio tokens, which leads to incorrect identification of the types and number of musical elements, potentially introducing duplication or omission errors. As a result, OmniSelect generates the correct answer, whereas OmniZip fails.

We further analyze cases where the Full Tokens setting produces the correct answer, while OmniSelect still maintains correct predictions after pruning 70\% of the tokens. As illustrated in Figure~\ref{fig:case3}, for the question \emph{``Which team, the white team or the black team, is more likely to win the entire game?''}, OmniSelect preserves substantially more tokens around critical frames that are highly relevant to the game dynamics, while aggressively masking redundant tokens in less informative regions. This selective retention effectively reduces irrelevant contextual interference during reasoning, enabling the model to maintain accurate predictions even under heavy compression. This observation also helps explain why, under the Any-Correct Evaluation setting in Table~\ref{tab:wb}, our method can even outperform the Full Tokens setting overall.

\section{Limitations and Future Work}

Although OmniSelect demonstrates that dynamically categorizing pruning strategies based on modality importance is effective for omni-modal token compression, our method still has several limitations. First, the current strategy selection mechanism relies on manually designed threshold-based rules, which may not fully capture the complex and continuously changing relationships between audio and visual modalities across different queries. As observed in our experiments, a single fixed threshold is often insufficient to optimally balance modality importance under diverse scenarios, leading to suboptimal pruning decisions in challenging cases. Second, our framework depends on AudioCLIP to estimate cross-modal relevance. While lightweight and efficient, AudioCLIP cannot always accurately distinguish fine-grained modality importance, especially when audio and visual semantics are highly entangled or weakly aligned. Future work could explore stronger yet lightweight modality-aware models specifically optimized for multimodal importance estimation and dynamic strategy prediction. Our evaluation is currently conducted within the Qwen2.5-Omni model family, using its 3B and 7B variants, and our primary comparison focuses on training-free compression methods. We include OmniZip as a matched baseline, but do not compare with OmniSIFT because it requires training with a straight-through estimator and its implementation was unavailable when our experiments were conducted. Thus, generalization to other omni-modal backbones and learned compression methods remains underexplored. Finally, although OmniSelect achieves competitive performance under aggressive compression ratios, there remains substantial room for further exploration in omni-modal token compression. Future research should investigate how to remove even more redundant multimodal tokens while preserving most of the reasoning capability and performance of OmniLLMs.


\end{document}